\documentclass{article} 
\usepackage{iclr2027_conference,times}
  \usepackage{amsmath,amssymb}

\usepackage{amsmath,amsfonts,bm}

\def\eqref#1{equation~\ref{#1}}

\def\1{\bm{1}}

\DeclareMathAlphabet{\mathsfit}{\encodingdefault}{\sfdefault}{m}{sl}
\SetMathAlphabet{\mathsfit}{bold}{\encodingdefault}{\sfdefault}{bx}{n}

\usepackage{hyperref}
\usepackage{url}
\usepackage{graphicx}
\usepackage{subcaption}
\usepackage{adjustbox}
\usepackage{booktabs}
\usepackage{multirow}
\usepackage{xcolor}

\usepackage{booktabs}
\usepackage{graphicx}
\usepackage{array}

\usepackage[most]{tcolorbox}

\title{NormTune: Norm-Guided Parameter-Efficient Fine-Tuning for Reasoning-Intensive Tasks}

\author{Zhangming Li\textsuperscript{1,2}, Qinghao Hu\textsuperscript{1,3}\thanks{Corresponding authors.},      Weifan Guan\textsuperscript{1,2}, Yifan Zhang\textsuperscript{4},  Peisong Wang\textsuperscript{1,2,3}, Jian Cheng\textsuperscript{1,2,3} \footnotemark[1]  \\
 \textsuperscript{1}The Key Laboratory of Cognition 
and Decision Intelligence for Complex Systems, 
\\ Institute of Automation, 
Chinese Academy of Sciences,\\
\textsuperscript{2}School of Artificial Intelligence, University of Chinese Academy of Sciences,\\
 \textsuperscript{3}AiRiA, \textsuperscript{4}School of Computer Science and Engineering, Nanjing University of Science and Technology\\
\texttt{\{lizhangming2023,huqinghao2014,guanweifan2024\}@ia.ac.cn}\\
\texttt{\{peisong.wang,jcheng\}@nlpr.ia.ac.cn, yifan\_zhang@njust.edu.cn}  
}

\iclrfinalcopy 
\begin{document}

\maketitle

\fancyhead{}
\renewcommand{\headrulewidth}{0pt}

\begin{abstract}

Full fine-tuning has achieved remarkable success on reasoning-intensive tasks, but it often requires substantial GPU resource consumption.
Parameter-efficient fine-tuning (PEFT) provides a promising alternative for reducing this cost, yet we observe a substantial performance gap on reasoning-intensive tasks. In this paper, we present three key observations showing that this gap is closely associated with norm-related under-adaptation, where PEFT produces markedly smaller weight-update norms than full fine-tuning.
Based on these observations, we propose Norm-Guided Parameter-Efficient Fine-Tuning (NormTune), a simple yet effective strategy that adaptively strengthens insufficiently updated layers according to their weight-update norms.
Extensive experiments show that NormTune consistently enhances multiple PEFT methods across challenging reasoning benchmarks.
Specifically, when trained on S1K-1.1, integrating NormTune into rank-8 LoRA, Bone, and MoRA yields substantial average accuracy improvements of 24.24\%, 11.33\%, and 16.21\% , respectively, across AIME 2024, MATH 500, GPQA, and AIME 2025. 
Moreover, NormTune-enhanced rank-256 LoRA, Bone, and MoRA achieve performance comparable to S1 full fine-tuning, with respective average gains of 2.17\%, 0.93\%, and 3.00\% over this baseline.
\end{abstract}

\section{Introduction}
\label{sec:introduction}

Recent studies \citep{Wu2025,OpenAI2024,DeepSeek2025,yu2025scaling,Xie2024} find that, for challenging reasoning tasks, carefully increasing inference-time compute can significantly boost performance. 
However, most of these approaches depend on substantial compute budgets, specialized training pipelines, or complex inference-time search, making them expensive, hard to reproduce, and impractical for broad adoption. To address these limitations, many studies\citep{Muennighoff2025,Ye2025,liu2025qfft,Yu2025Z1} have explored simpler and more reproducible alternatives based on supervised fine-tuning (SFT). 
These methods, however, still depend on full fine-tuning, which becomes progressively less practical as model scale increases, particularly in compute-constrained settings. 
For example, full fine-tuning of Qwen2.5-32B-Instruct \citep{qwen2.5} on a dataset with a 32K context length requires 562.07 GB of memory even with gradient checkpointing, making it difficult to apply under limited compute resources.
To address this challenge, we consider parameter-efficient fine-tuning (PEFT) methods \citep{hu2021lora,lin2024nora,feng2025omoe}, which adapt large pretrained models to downstream tasks by updating only a small subset of parameters while keeping the backbone model frozen. 

\begin{figure*}[t]
  \centering

  \begin{subfigure}[t]{0.36\textwidth}
    \centering
    \includegraphics[width=\linewidth]{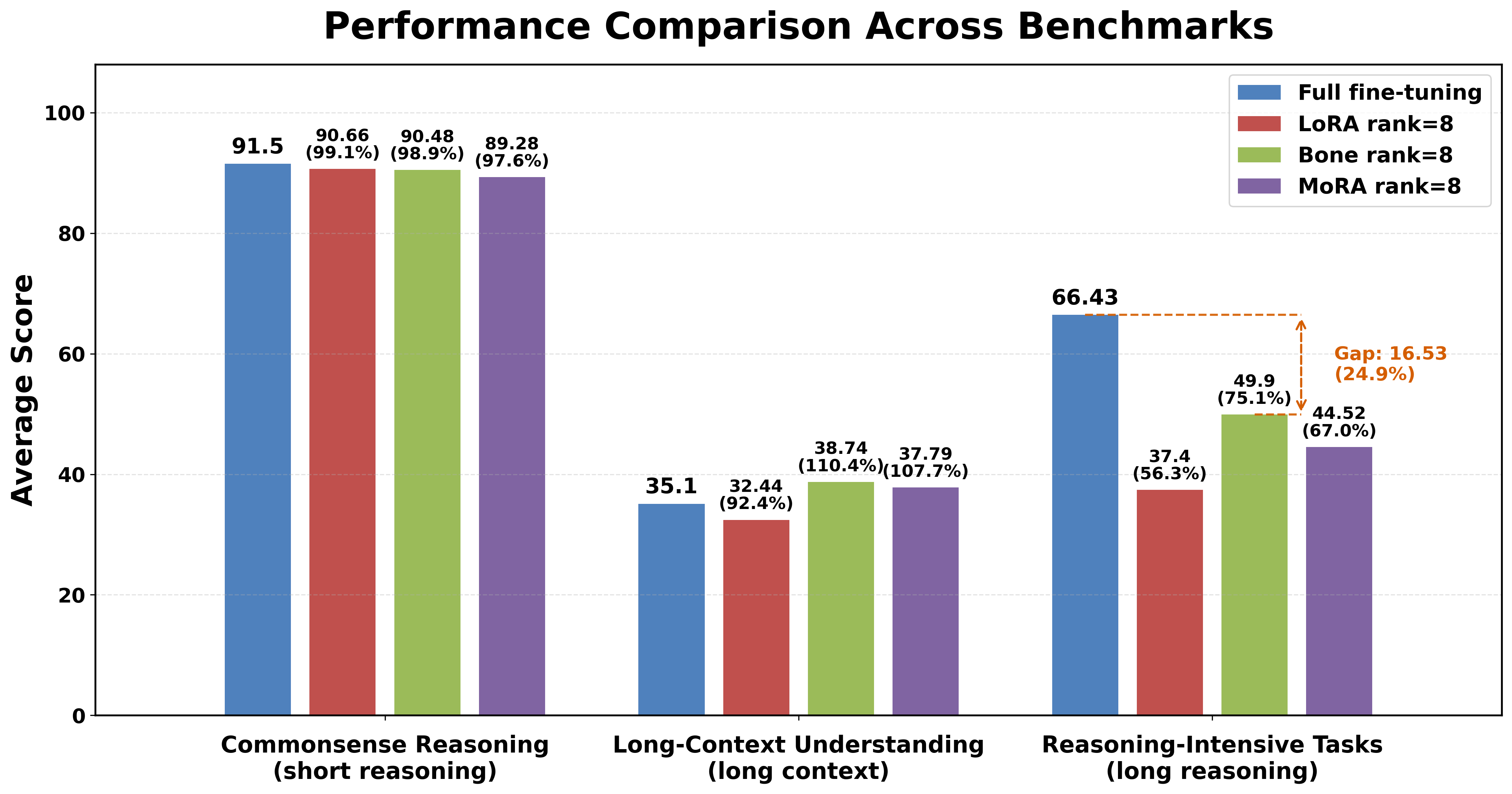}
    \caption{Performance comparison across evaluation regimes. PEFT methods exhibit a substantial gap from full fine-tuning only on reasoning-intensive tasks using Qwen2.5-32B-Instruct.}
    \label{fig:tasks_compare}
  \end{subfigure}
  \hspace{0.04\textwidth}
  \begin{subfigure}[t]{0.36\textwidth}
    \centering
    \includegraphics[width=\linewidth]{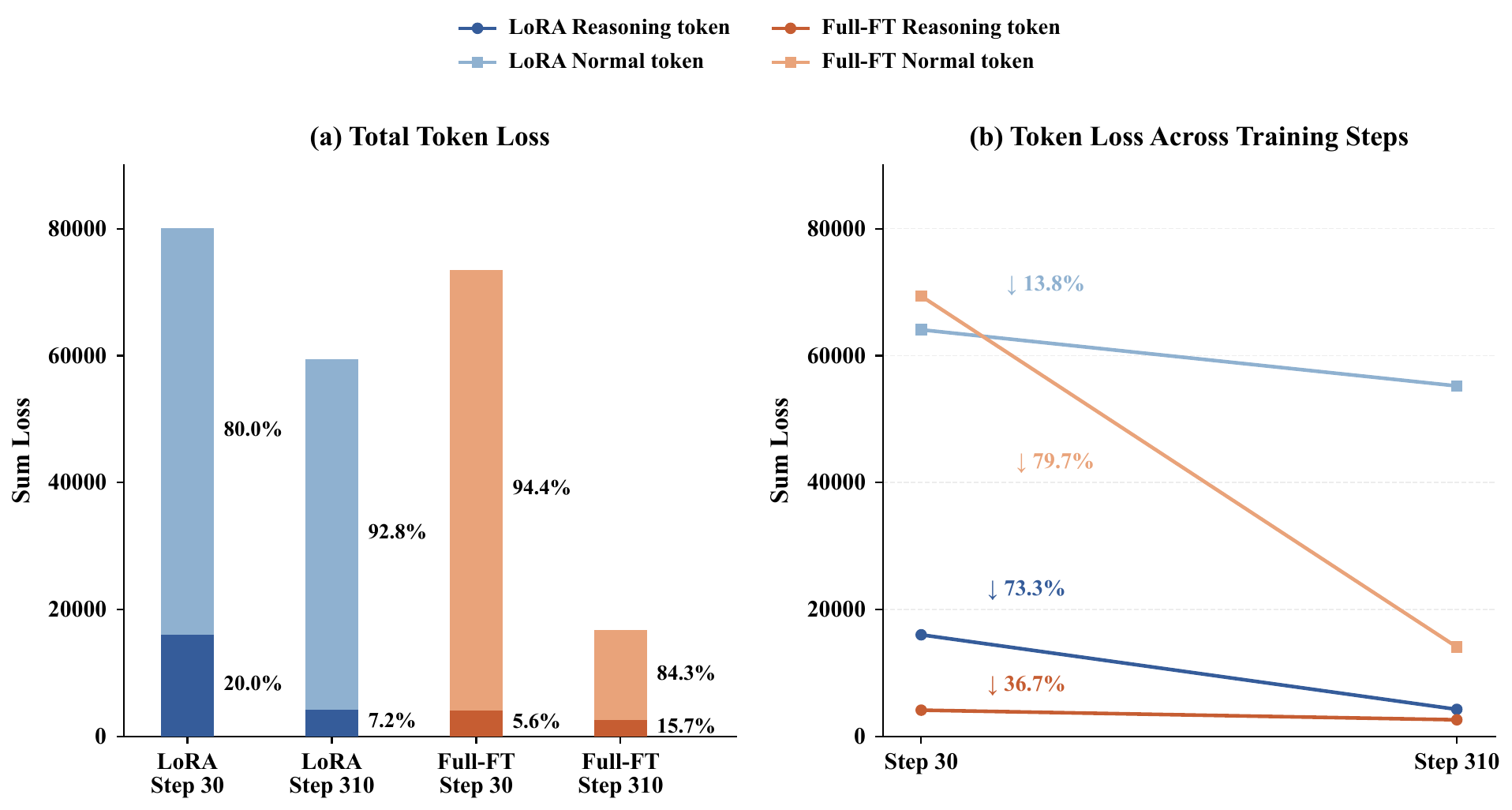}
    \caption{Comparison of token-level and total loss values between rank-256 LoRA and full fine-tuning on 16 identical sequences from reasoning-intensive tasks using Qwen2.5-32B-Instruct.}
    \label{fig:tokenloss}
  \end{subfigure}

  \caption{
  Performance across evaluation regimes and token-level loss under PEFT and full fine-tuning. Detailed results are reported in Appendix Tables~\ref{commonsense-table}, \ref{government_table}, and \ref{s1_table}.
  }
  \label{fig:memory_task_compare}
\end{figure*}

Motivated by this efficiency, we apply PEFT to reasoning-intensive adaptation, while noting that its reduced training cost may come with non-negligible performance risks in such challenging settings. Thus we investigate the question: \textbf{Can low-rank PEFT retain the Full-FT comparable adaptation capability?} As shown on the right side of Figure \ref{fig:tasks_compare}, existing PEFT methods exhibit a substantial performance drop on reasoning-intensive tasks, failing to reach 90 \% of the full fine-tuning performance commonly used criterion for effective low-dimensional adaptation \citep{aghajanyan2021intrinsic}. In contrast, PEFT remains effective on tasks involving shorter reasoning chains and long-context understanding, as shown in the left and middle panels of Figure~\ref{fig:tasks_compare}.  
To examine this gap, we present three key observations. 
\textbf{Observation 1:} PEFT methods exhibit a much larger loss contribution on reasoning tokens such as transition, self-verification, template, and reasoning-boundary tokens compared with Full-FT at earlier training. During training, PEFT methods exhibit a smaller relative reduction in loss on normal tokens (i.e., all tokens other than reasoning tokens) than on reasoning tokens, whereas Full-FT exhibits the opposite pattern, as shown in Figure~\ref{fig:tokenloss}. 
\textbf{Observation 2:} The gradients induced by reasoning token and normal token losses exhibit substantial conflicts across most trainable parameters in earlier layers. 
However, the total token gradients remain aligned with the normal token optimization direction, while its effective magnitude is partially reduced by the conflicting reasoning token gradients. This gradient interference weakens the effective parameter updates in earlier layers.
\textbf{Observation 3:} PEFT methods induce substantially smaller weight-update norms than full fine-tuning, especially in earlier layers.  
To achieve performance comparable to Full-FT, we expect PEFT methods to exhibit similar layer-wise weight-update magnitudes as Full-FT. Based on this observation, we propose Norm-Guided Parameter-Efficient Fine-Tuning (NormTune), which reduces the gradients conflicts in earlier layers, reallocates optimization strength toward under-adapted layers and improves model performance. 

Our contributions can be summarized as follows:
\begin{itemize}

   \item We systematically investigate the limitations of PEFT on reasoning-intensive tasks and present three key observation, revealing that PEFT induces substantially smaller fine-tuning-induced weight changes, which is closely associated with its performance degradation.
    \item Motivated by these observations, we propose a novel PEFT method named  NormTune, which reduces the gradients conflicts in earlier layers and reallocates optimization strength toward under-adapted layers. To the best of our knowledge, NormTune is the first approach that enables PEFT methods to achieve comparable performance relative to full fine-tuning on reasoning-intensive tasks.
    \item Extensive experiments demonstrate the effectiveness and generality of NormTune. Specifically, when trained on S1K-1.1, NormTune improves rank-8 LoRA, Bone, and MoRA by 24.24\%, 11.33\%, and 16.21\% in average accuracy across AIME 2024, MATH 500, GPQA, and AIME 2025, respectively. Moreover, NormTune-enhanced rank-256 LoRA, Bone, and MoRA achieve performance comparable to S1 full fine-tuning, with respective average gains of 2.17\%, 0.93\%, and 3.00\% over this baseline.
\end{itemize}

\begin{figure*}[t]
  \centering
  \begin{subfigure}[t]{0.32\textwidth}
    \centering
    \includegraphics[width=\linewidth]{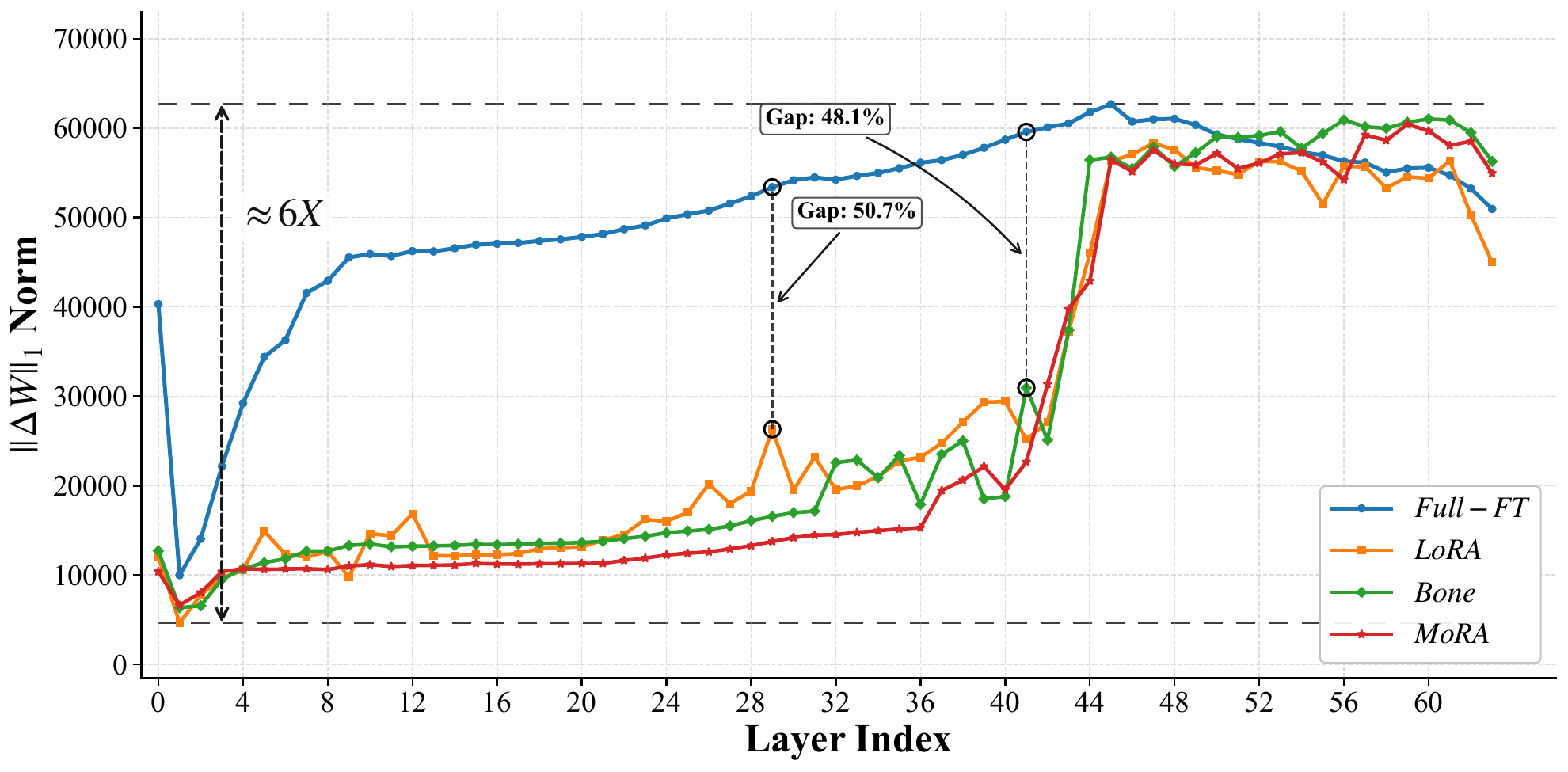}
    \caption{Visualization of entry-wise weight-update norms of full fine-tuning and PEFT methods}
    \label{fig:vs_total_4b}
  \end{subfigure}
  \hfill
  \begin{subfigure}[t]{0.32\textwidth}
    \centering
    \includegraphics[width=\linewidth]{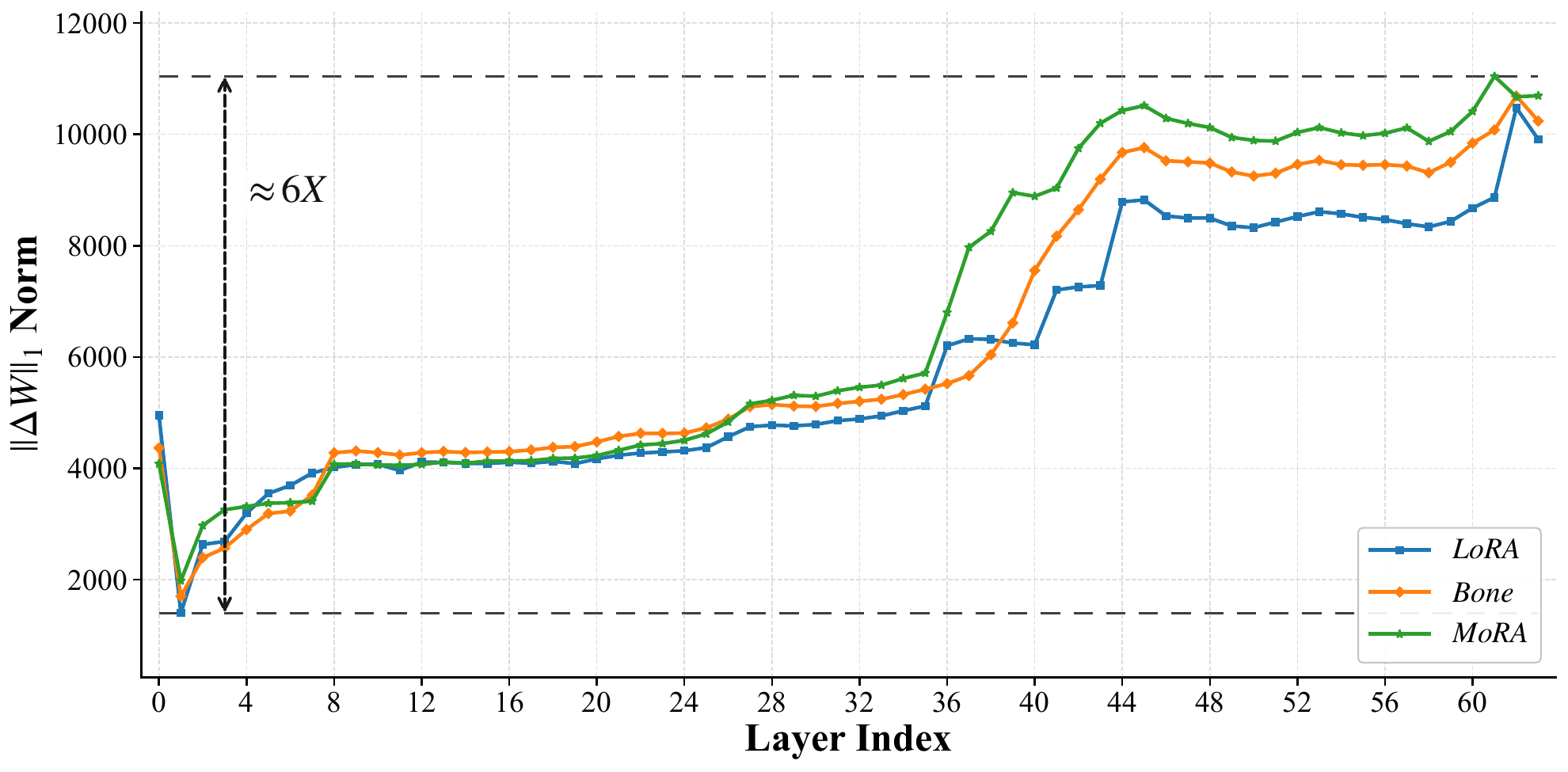}
    \caption{Visualization of weight-update norm of different PEFT methods after 30 steps training}
    \label{fig:vs_total_7b}
  \end{subfigure}
  \hfill
  \begin{subfigure}[t]{0.32\textwidth}
    \centering
    \includegraphics[width=\linewidth]{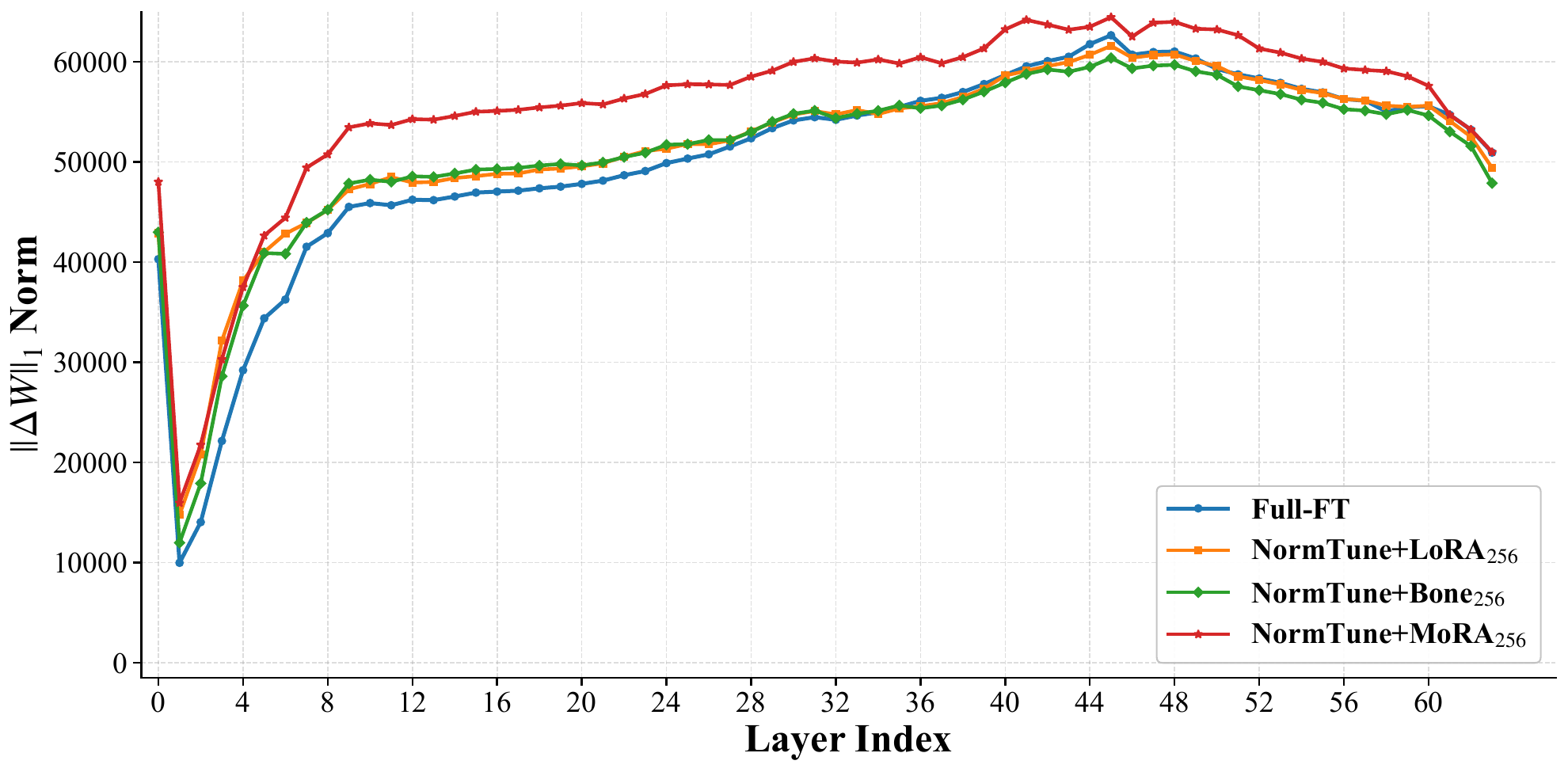}
    \caption{Visualization of entry-wise weight-update norms of full fine-tuning and NormTune-enhanced PEFT methods}
    \label{fig:peft_our_norm}
  \end{subfigure}

  \caption{Layer-wise norm comparison among Full-FT, rank-256 PEFT methods, early-stage PEFT statistics, and NormTune+PEFT methods.}
  \label{fig:vs_total}
\end{figure*}

\begin{figure*}[t]
  \centering
  \begin{subfigure}[t]{0.32\textwidth}
    \centering
    \includegraphics[width=\linewidth]{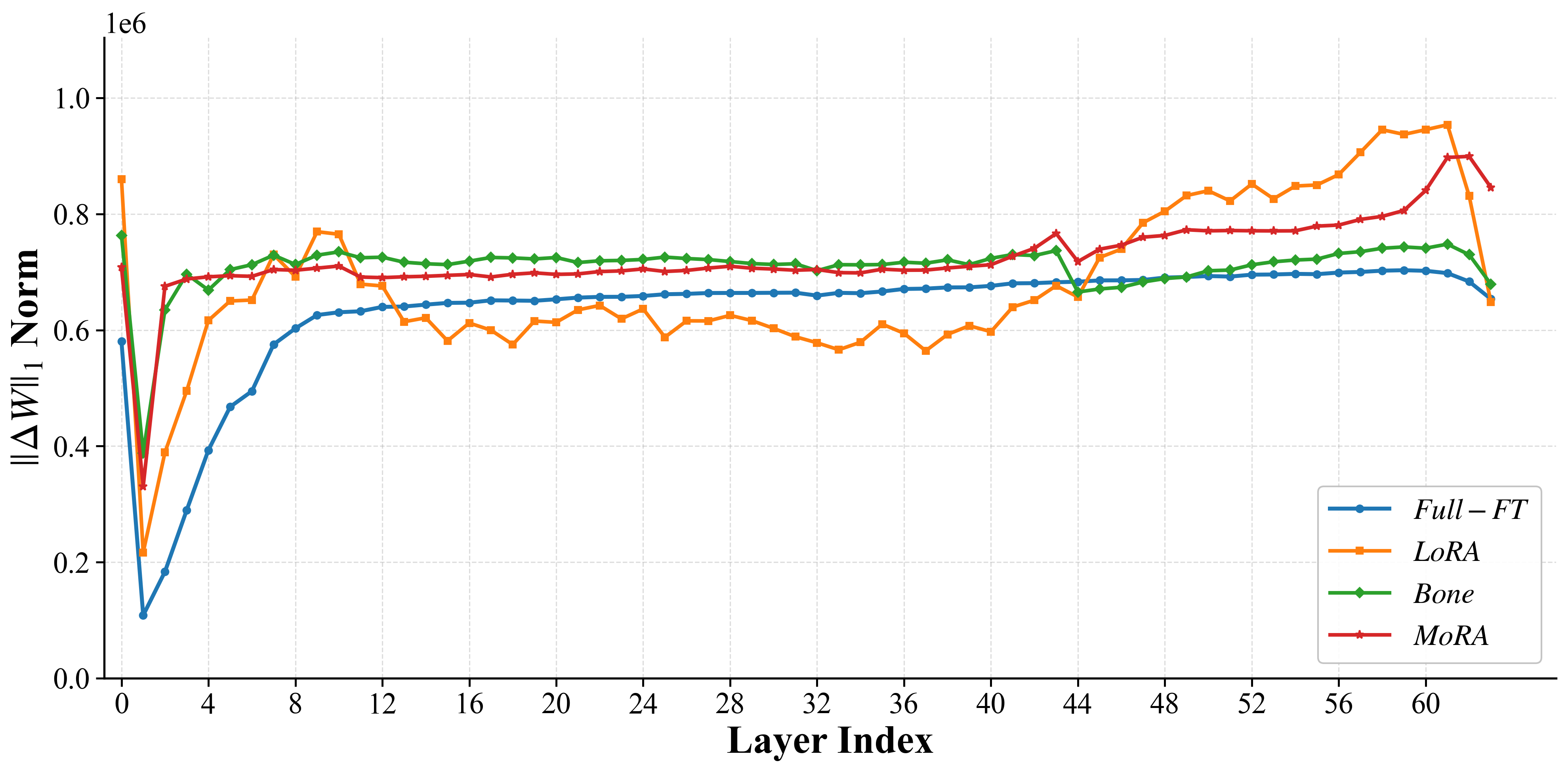}
    \caption{Commonsense reasoning task}
    \label{fig:commmon}
  \end{subfigure}
  \hfill
  \begin{subfigure}[t]{0.32\textwidth}
    \centering
    \includegraphics[width=\linewidth]{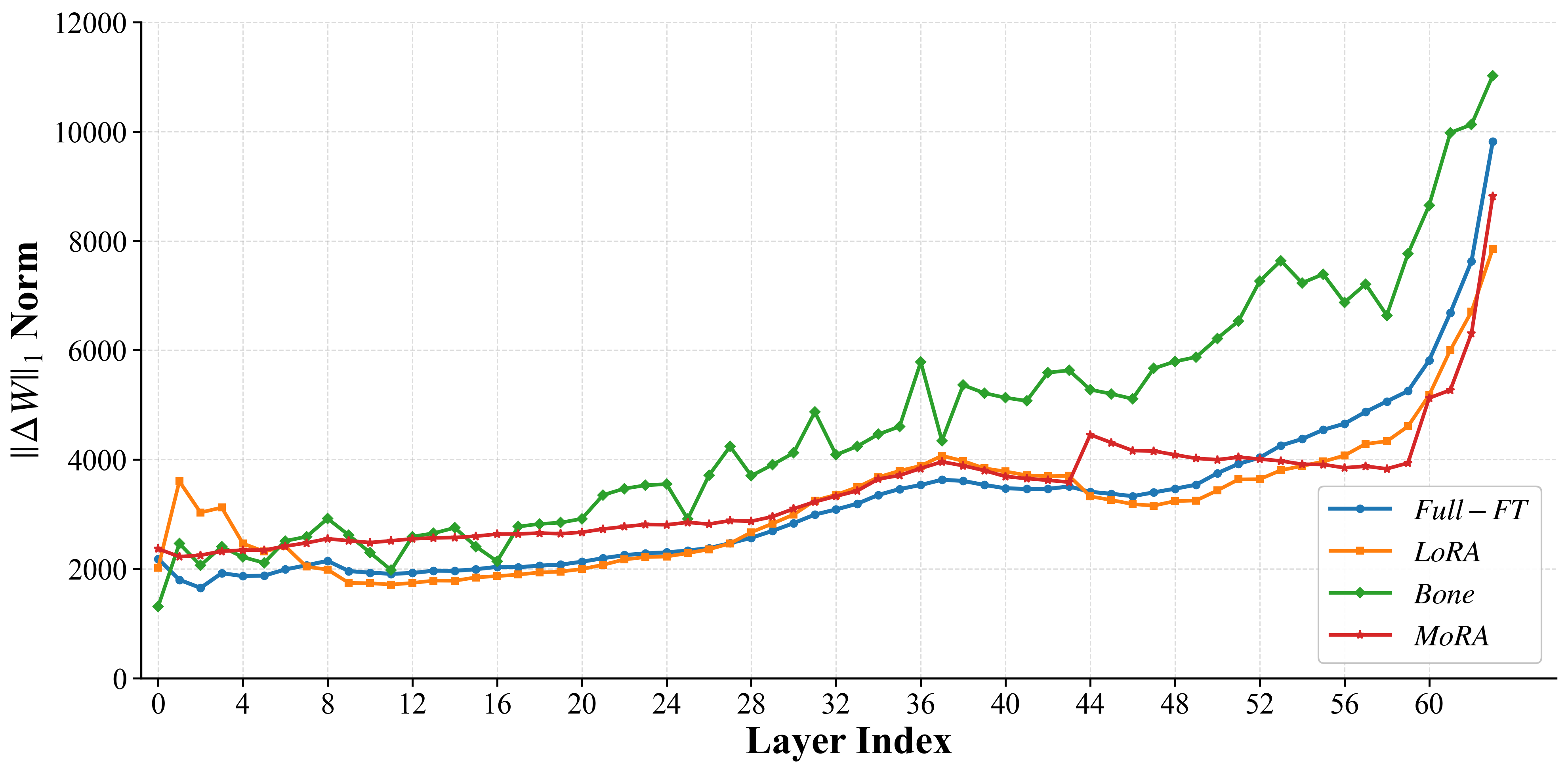}
    \caption{GovReport task}
    \label{fig:government}
  \end{subfigure}
  \hfill
  \begin{subfigure}[t]{0.32\textwidth}
    \centering
    \includegraphics[width=\linewidth]{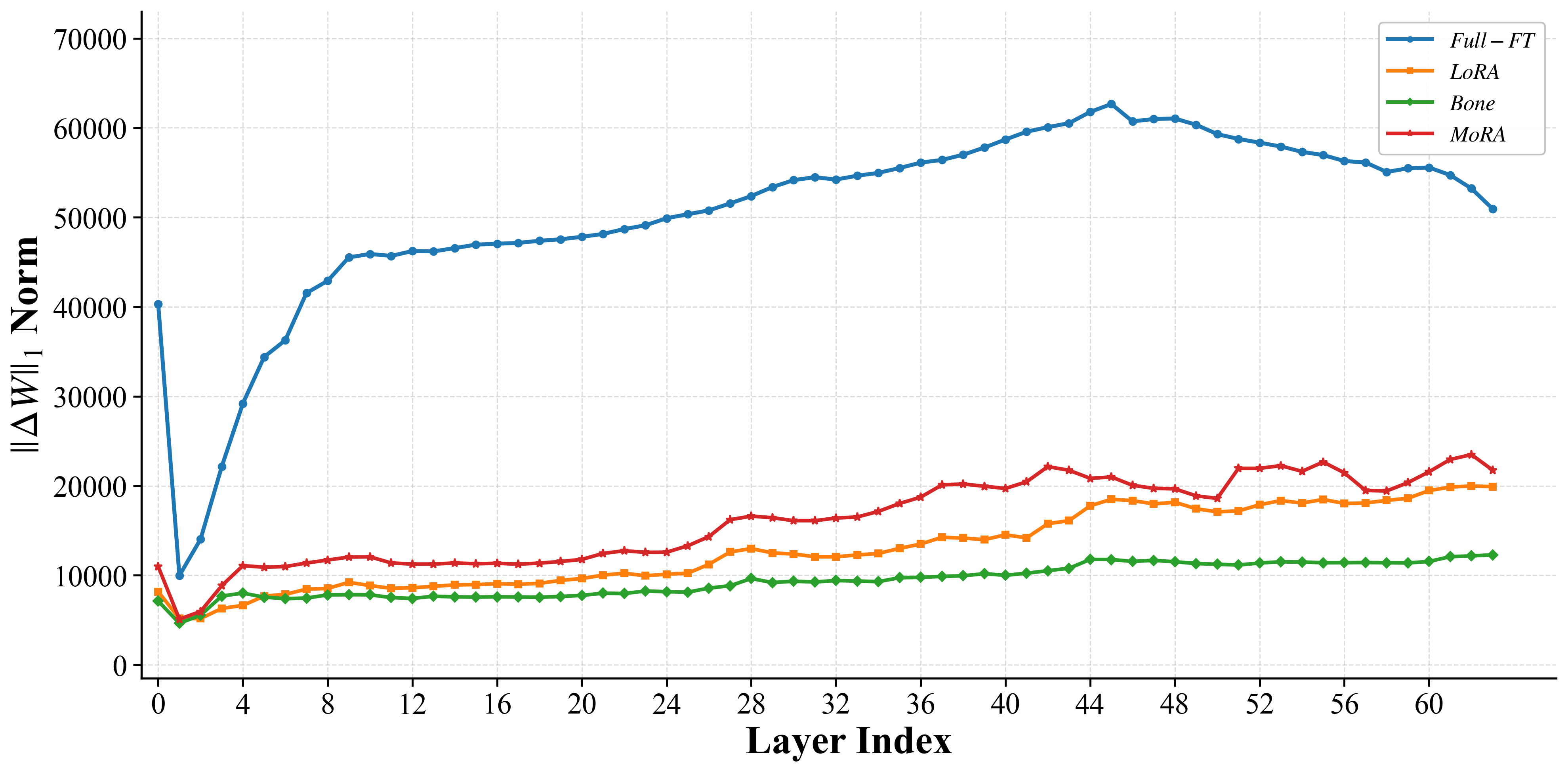}
    \caption{Reasoning-intensive task}
    \label{fig:s1peftrank8}
  \end{subfigure}
  \hfill
  
  \caption{
  Layer-wise norm comparison between full fine-tuning (Full-FT) and rank-8 PEFT methods on (a) commonsense reasoning, (b) GovReport, and (c) reasoning-intensive tasks.}
  
  \label{fig:common_gover}
\end{figure*}

  


\section{Methodology}
\label{sec:method}


          
          

\subsection{Key Observations On Reasoning-Intensive Tasks}

\textbf{Observation 1:  PEFT methods exhibit a much larger loss contribution on reasoning tokens compared with Full-FT at earlier training.}
To investigate the accuracy gap between PEFT and Full-FT on reasoning-intensive tasks, we analyze how training loss is distributed across tokens and how it evolves during fine-tuning, motivated by prior work \cite{2024Rho,2025Token}. 
Specifically, we sample 16 sequences from S1K-1.1~\citep{Muennighoff2025} and evaluate token-level cross-entropy losses on the same sequences at training step 30, immediately after warm-up, and step 310, at the end of training, using Qwen2.5-32B-Instruct.
We inspect the high-loss tail using the empirical 99th percentile ($P_{99}$)\cite{sato2022anomaly} of token-level losses as a cutoff. Most tokens above this threshold correspond to transitions, self-verification expressions, templates, or reasoning boundaries such as ``Alternatively'' and ``Wait''.. For this analysis, we refer to tokens serving these four roles as reasoning tokens, and to all remaining tokens as normal tokens. As shown in Figure~\ref{fig:tokenloss}, PEFT methods exhibit a much larger loss contribution on reasoning tokens compared with Full-FT at earlier training. After training at step 310 , PEFT methods exhibit a smaller relative reduction in loss on normal tokens than on reasoning tokens, whereas Full-FT exhibits the opposite pattern. This indicates that PEFT disproportionately allocates its optimization capacity toward adapting reasoning token patterns, while providing insufficient adaptation for normal token representations compared with Full-FT.

\textbf{Observation 2:  The gradients induced by reasoning token and normal token losses exhibit substantial conflicts across most trainable parameters in earlier layers.} 

\begin{figure*}[t]
    \centering

    \begin{subfigure}[c]{0.38\textwidth}
        \centering
        \includegraphics[width=\linewidth]{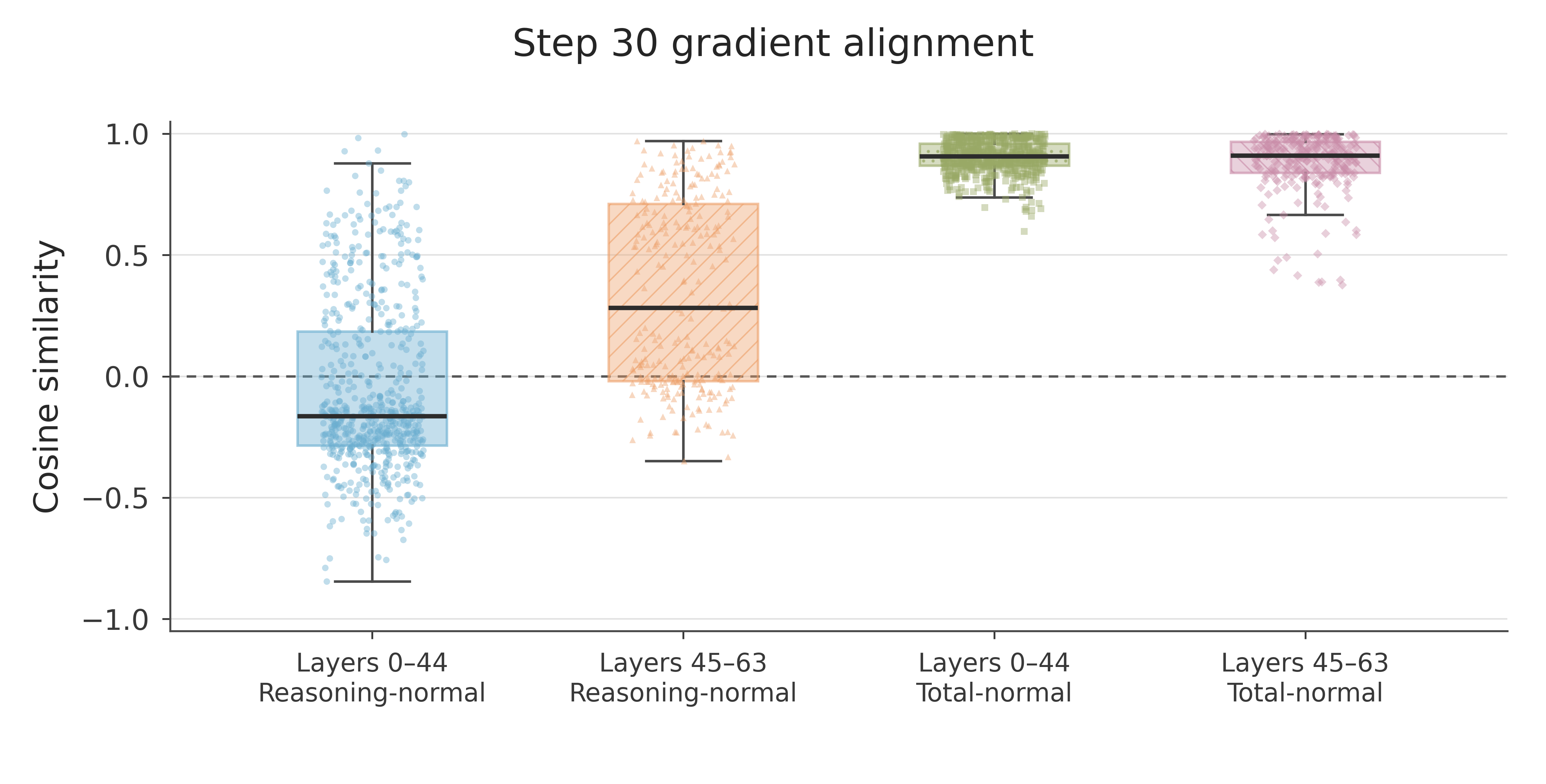}
        \caption{
    Gradient alignment across early and late layers, measured by
    $\mathrm{CosSim}(\nabla_{\theta}\mathcal{L}_{\mathrm{reasoning}},
    \nabla_{\theta}\mathcal{L}_{\mathrm{normal}})$ and
    $\mathrm{CosSim}(\nabla_{\theta}\mathcal{L}_{\mathrm{total}},
    \nabla_{\theta}\mathcal{L}_{\mathrm{normal}})$. The corresponding Full-FT results are provided in Figure~\ref{fig:full_xiang}.
}
        \label{fig:lora_AB_cosine}
    \end{subfigure}
    \hfill
    %
    \begin{subfigure}[c]{0.58\textwidth}
        \centering
        
        \begin{subfigure}[c]{0.48\linewidth}
            \centering
            \includegraphics[width=\linewidth]{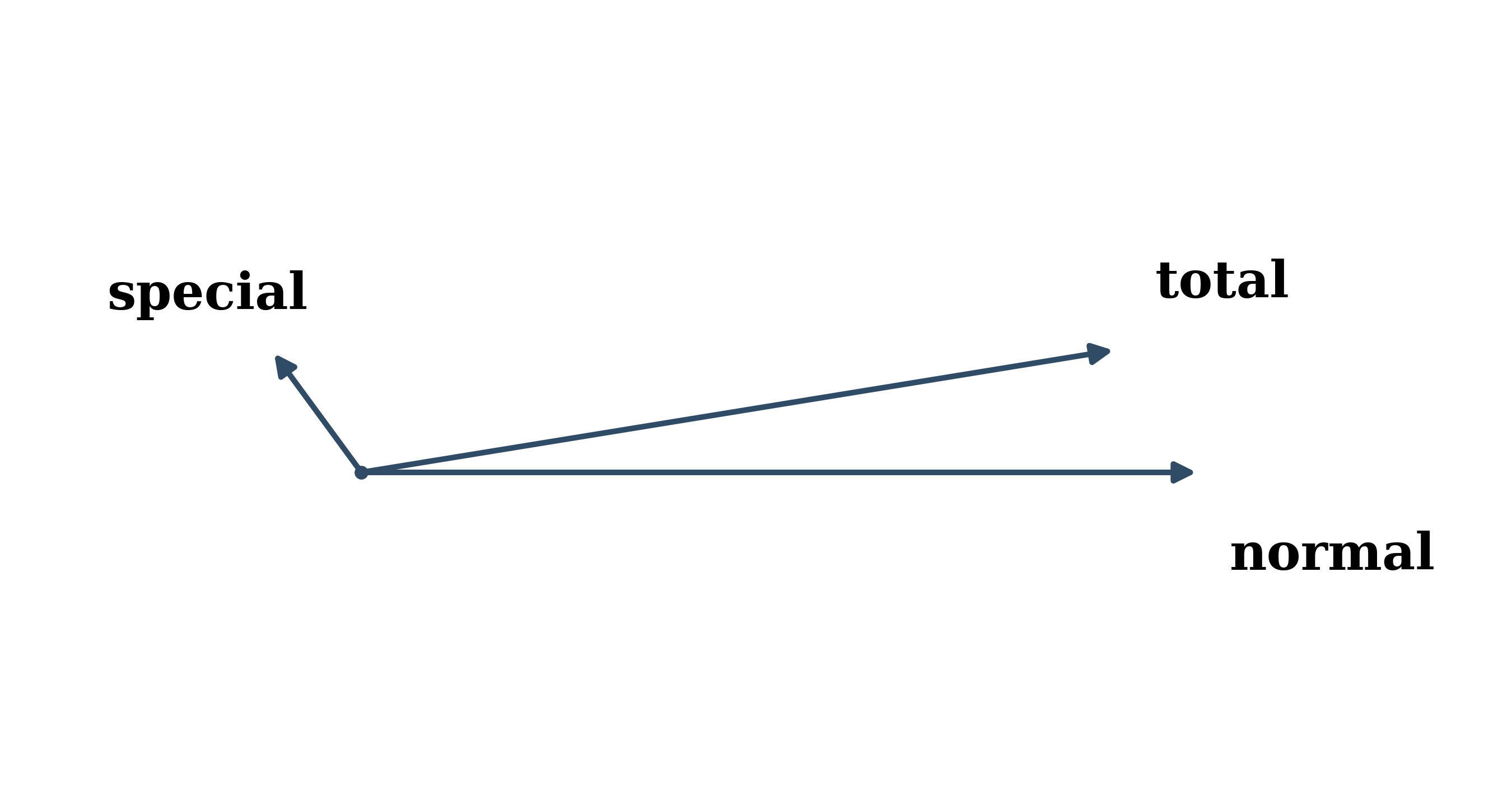}
            \caption{Layer 4 MLP down ($A$)}
            \label{fig:layer4_down_A}
        \end{subfigure}
        \hfill
        \begin{subfigure}[c]{0.48\linewidth}
            \centering
            \includegraphics[width=\linewidth]{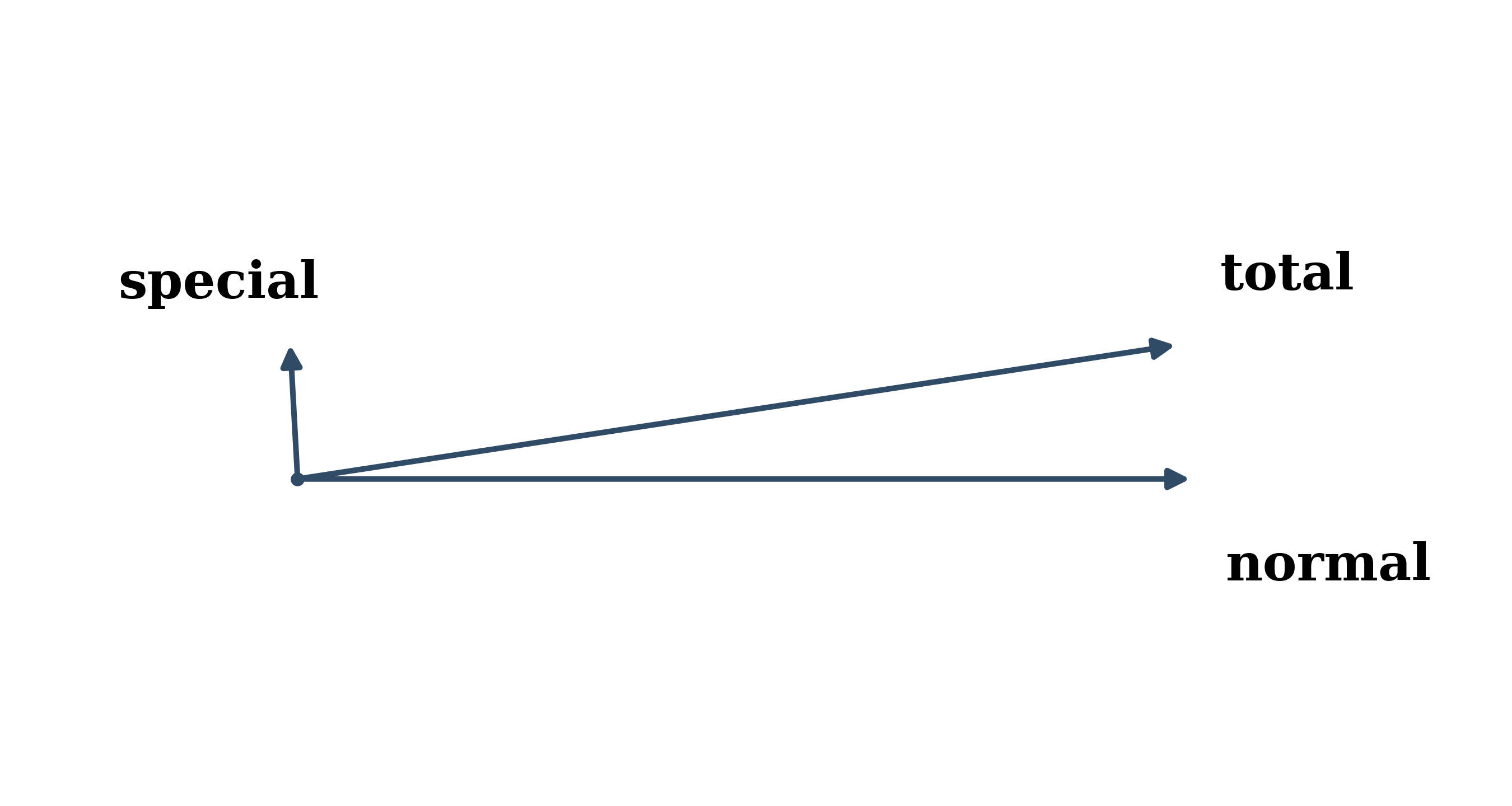}
            \caption{Layer 41 attention Q ($B$) }
            \label{fig:layer41_v_B}
        \end{subfigure}

        \vspace{0.5em}

        \begin{subfigure}[c]{0.48\linewidth}
            \centering
            \includegraphics[width=\linewidth]{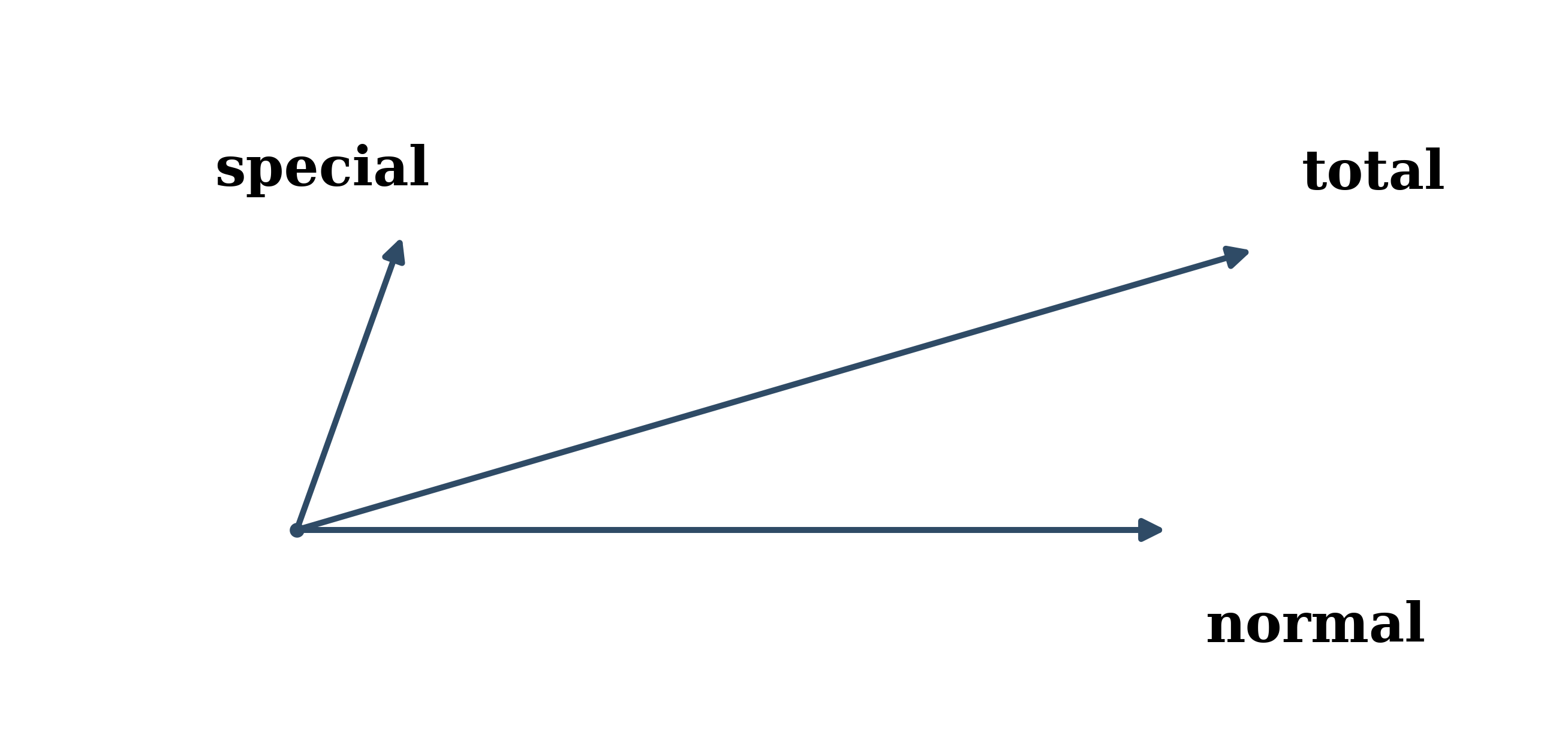}
            \caption{ Layer 53 attention V ($A$)}
            \label{fig:layer53_v_A}
        \end{subfigure}
        \hfill
        \begin{subfigure}[c]{0.48\linewidth}
            \centering
            \includegraphics[width=\linewidth]{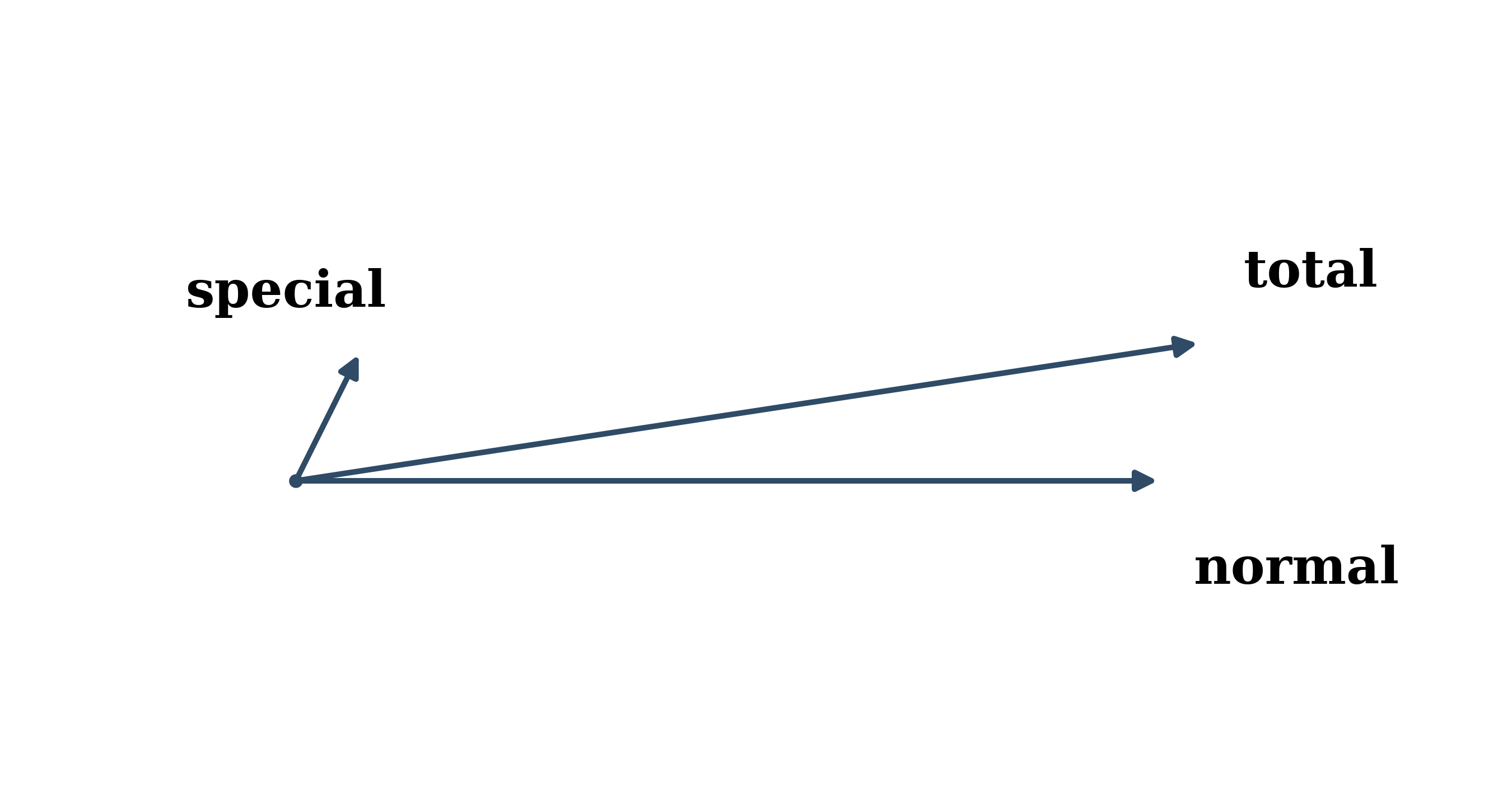}
            \caption{ Layer 57 MLP gate ($B$)}
            \label{fig:layer57_gate_A}
        \end{subfigure}

    \end{subfigure}

    \caption{
        Gradient alignment analysis between token groups and layer-wise optimization directions.
        The left panel shows the cosine similarity between different gradients.
       The right panels visualize representative gradient directions and the L1 norms of the gradients for different LoRA parameters (LoRA A and B) across early and late layers after 30 training steps.
    }
    \label{fig:gradient_direction_analysis}
\end{figure*}

To understand how different token groups affect the optimization dynamics of PEFT methods, we decompose the training loss into reasoning token and normal token components.
Let $\theta$ denote all trainable LoRA parameters. 
The total loss can be rewritten as
\begin{align}
    \mathcal{L}_{\mathrm{total}}(\theta)
    &=
    \frac{1}{|\mathcal{I}|}
    \left(
    \sum_{(i,t)\in\mathcal{I}_{\mathrm{reasoning}}}
    \ell_{i,t}(\theta)
    +
    \sum_{(i,t)\in\mathcal{I}_{\mathrm{normal}}}
    \ell_{i,t}(\theta)
    \right)
    =
    \mathcal{L}_{\mathrm{reasoning}}(\theta)
    +
    \mathcal{L}_{\mathrm{normal}}(\theta),
    \label{eq:loss_decomposition}
\end{align}
where $\mathcal{I}$ denote total tokens, $i$ is sequence and the $t$ is prediction position.
Because both loss terms are differentiable with respect to the LoRA parameters $\theta$, taking the gradient of Eq.~\ref{eq:loss_decomposition} yields $\nabla_{\theta}\mathcal{L}_{\mathrm{total}} = \nabla_{\theta}\left(\mathcal{L}_{\mathrm{special}} + \mathcal{L}_{\mathrm{normal}}\right) = \nabla_{\theta}\mathcal{L}_{\mathrm{special}} + \nabla_{\theta}\mathcal{L}_{\mathrm{normal}}.$
Therefore, the update direction of LoRA parameters is determined by the combination of two independent optimization signals:
the gradient induced by reasoning tokens and the gradient induced by normal tokens.
We define their gradient alignment using cosine similarity. Specifically, for 
$\nabla_{\theta}\mathcal{L}_{\mathrm{special}}$ and $\nabla_{\theta}\mathcal{L}_{\mathrm{normal}}$, the cosine similarity is defined as
$\mathrm{CosSim}(\nabla_{\theta}\mathcal{L}_{\mathrm{special}},\nabla_{\theta}\mathcal{L}_{\mathrm{normal}})=
\frac{\nabla_{\theta}\mathcal{L}_{\mathrm{special}}^{\top}\nabla_{\theta}\mathcal{L}_{\mathrm{normal}}}
{\|\nabla_{\theta}\mathcal{L}_{\mathrm{special}}\|_2\|\nabla_{\theta}\mathcal{L}_{\mathrm{normal}}\|_2}$.
We also measure their alignment with the overall optimization direction
$\mathrm{CosSim}(\nabla_{\theta}\mathcal{L}_{\mathrm{special}},
\nabla_{\theta}\mathcal{L}_{\mathrm{total}})$ and
$\mathrm{CosSim}(\nabla_{\theta}\mathcal{L}_{\mathrm{normal}},
\nabla_{\theta}\mathcal{L}_{\mathrm{total}})$.
As shown in Figure~\ref{fig:gradient_direction_analysis}, the gradients induced by special-token and normal-token losses exhibit substantial conflicts across most trainable parameters in earlier layers. 
However, this behavior is reversed in deeper layers, where the two optimization signals become more compatible. 
Although earlier layers exhibit
conflicts between the two token-group gradients, the total gradient remains positively aligned with the normal-token gradient. This
indicates that the special-token gradient does not reverse the
normal-token optimization direction, but reduces its effective update magnitude. 

\textbf{Observation 3: PEFT methods induce substantially smaller  weight-update norms than full fine-tuning, especially in earlier layers.} 


Motivated by norm-based generalization theory~\cite{bartlett2017spectrally, galanti2023norm, nagarajan2019generalization}, which links network norm to model capacity and generalization, we analyze the magnitude of fine-tuning-induced weight changes as a proxy for layer-wise adaptation strength. 
Specifically, for module $m$ in layer $l$, we define the parameter update under fine-tuning strategy $\tau$ as
$\Delta W_{l,m}^{\tau} = W_{l,m}^{\tau} - W_{l,m}^{0}$,
where $W_{l,m}^{0}$ denotes the corresponding base-model weight and $W_{l,m}^{\tau}$ denotes the fine-tuned weight. Here, $\tau \in \{\mathrm{Full}, \mathrm{PEFT}\}$, where $\mathrm{Full}$ represents full fine-tuning and $\mathrm{PEFT} \in \{\mathrm{LoRA}, \mathrm{Bone}, \mathrm{MoRA}\}$.
To quantify the adaptation strength of each layer, we aggregate the entry-wise $L_1$ norm of the effective weight updates over all target modules in that layer:
\begin{equation}
U_l^{\tau} =
    \sum_{m \in \mathcal{M}_l}
    \left\|
    \Delta W_{l,m}^{\tau}
    \right\|_{1,1}, \space
 where
    \left\|
    \Delta W_{l,m}^{\tau}
    \right\|_{1,1}
    =
    \sum_{i,j}
    \left|
    \Delta W_{l,m,ij}^{\tau}
    \right|.
\end{equation}

 Here, $\mathcal{M}_l$ denotes the set of target modules in layer $l$, including the attention projections $\{Q,K,V,O\}$ and the MLP projections $\{\mathrm{Gate}, \mathrm{Up}, \mathrm{Down}\}$. In addition, the LoRA scaling factor $s=\alpha/r=2$, which is already incorporated in this function.
Based on this metric, Figure~\ref{fig:vs_total_4b} shows that PEFT methods with rank 256 induce substantially smaller weight-update norms than full fine-tuning, with the gap being particularly pronounced in earlier layers. This indicates that PEFT provides much weaker effective adaptation in these layers.
Furthermore, we visualize the layer-wise weight-update norms of rank-8 PEFT methods on other tasks, as shown in Figure~\ref{fig:common_gover}. Unlike reasoning-intensive tasks, these benchmarks do not exhibit significant under-adaptation in earlier layers, suggesting that the insufficient layer adaptation observed in PEFT is particularly associated with reasoning-intensive fine-tuning.

\subsection{NormTune}

Overall, the three observations suggest that improving PEFT on reasoning-intensive tasks requires stronger adaptation for normal tokens, particularly in earlier layers. Observation~1 reveals an imbalance in token-level loss reduction, suggesting insufficient adaptation for normal tokens relative to Full-FT. Observation~2 provides a explanation: although reasoning-token and normal-token gradients conflict in earlier layers, the total gradient remains positively aligned with the normal-token gradient. This interference therefore attenuates the gradient component along the normal-token direction while preserving its positive alignment. This pattern is consistent with Observation~3, which shows substantially smaller weight-update norms in these layers under PEFT. 
These findings motivate increasing the learning rates of under-adapted layers to compensate for attenuated updates. 
Specially, for fixed parameters, gradients, and optimizer state, increasing a layer's learning rate scales the magnitude of its parameter update without changing its direction.



Motivated by these observations,
we propose Norm-Guided Parameter-Efficient Fine-Tuning (NormTune), which assigns larger learning rates to weakly updated layers based on their weight-update norms, thereby improving the adaptation of under-trained layers.
In general, $U_l^{\mathrm{Full}}$, i.e., the layer-wise weight-update norm obtained from full fine-tuning, can serve as an ideal reference scale for calibrating PEFT learning, encouraging $U_l^{\mathrm{PEFT}}$ to align with the adaptation magnitude of full fine-tuning.  However, obtaining $U_l^{\mathrm{Full}}$ requires performing full fine-tuning, which is impractical due to its substantial training time and computational cost. Here,we first examine whether the layer-wise norm statistics from PEFT can serve as a practical alternative. Empirically, we observe that after a complete PEFT run, the maximum-to-minimum ratio of layer-wise update norms, $\max_l U_l^{\mathrm{PEFT}} / \min_l U_l^{\mathrm{PEFT}}$, is close to that of full fine-tuning, with both ratios being approximately $6\times$. This indicates that although PEFT produces much smaller absolute update norms, it still captures a similar relative range of layer-wise adaptation magnitudes.

Nevertheless, obtaining $U_l^{\mathrm{PEFT}}$ from a completed PEFT run remains costly. We therefore introduce a lightweight probing procedure that performs only 30 PEFT training steps. As shown in Figure~\ref{fig:vs_total_7b}, the early-stage update norms $U_l^{\mathrm{PEFT},s}$ already exhibit a comparable maximum-to-minimum ratio of approximately $6\times$. Based on this observation, we adopt this early-stage statistic as an efficient proxy for estimating the desired layer-wise adaptation range.
Directly assigning a distinct learning rate to each layer may be sensitive to local fluctuations in $U_l^{\mathrm{PEFT},s}$. 
We therefore group layers into $n$ buckets according to their update magnitudes inspired by \cite{zhang2025adam}. 
Let $U_{\min}^{\mathrm{PEFT},s}=\min_l  U_{l}^{\mathrm{PEFT},s}$, $U_{\max}^{\mathrm{PEFT},s}=\max_l U_{l}^{\mathrm{PEFT},s}$.
The range $[U_{\min}^{\mathrm{PEFT},s},U_{\max}^{\mathrm{PEFT},s}]$ is uniformly divided into $n$ buckets. 
For the $k$-th bucket, we denote by $\bar{U}_{k}^{\mathrm{PEFT},s}$ the mean update magnitude of the layers inside this bucket. 
Following the standard min-max normalization used in data preprocessing,
which rescales numerical values into a fixed range while preserving their relative order~\cite{mining2006data}, we normalize
$\bar{U}_{k}^{\mathrm{PEFT},s}$ as
\begin{equation}
    q_k =
    \frac{\bar{U}_{k}^{\mathrm{PEFT},s}-\bar{U}_{min}^{\mathrm{PEFT},s}}
    {\bar{U}_{max}^{\mathrm{PEFT},s}-\bar{U}_{min}^{\mathrm{PEFT},s}},
    \qquad
    q_k \in [0,1].
    \label{eq:NormTune_normalized_score}
\end{equation}
Here, $q_k$ measures the relative adaptation sufficiency of bucket $k$.
We define the bucket-wise learning rate by mapping $q_k$ to the interval $[\eta_{\min},\eta_{\max}]$:
\begin{equation}
    \eta_k
    =
    \eta_{\min}
    +
    (\eta_{\max}-\eta_{\min}) h(q_k),
    \label{eq:NormTune_general_mappingd}
\end{equation}
where $\eta_{\min}$ denotes the base learning rate, $\eta_{\max} > \eta_{\min}$.
The function $h$ should satisfy three simple requirements.
First, the weakest adapted bucket should receive the largest learning rate, and the strongest adapted bucket should receive the smallest learning rate:$    h(0)=1, h(1)=0.$
Second, the learning rate should decrease as the adaptation magnitude increases, so $h(q)$ should be monotonic non-increasing on $[0,1]$. 
Third, buckets near $q=0$ should remain close to the maximum learning rate, because they all correspond to insufficiently adapted layers. 
We encode this weak-adaptation smoothing condition as 
$ h'(0)=0$.
To keep the rule simple, we choose the lowest-degree polynomial satisfying these constraints. 
Let $ h(q)=a+bq+cq^2$. Given $h(0)=1$, $h'(0)=0$, and $h(1)=0$, the solution of $h(q)$ is: 
\begin{equation}
    h(q)=1-q^2 .
    \label{eq:NormTune_quadratic_scored}
\end{equation}
Substituting Eq.~\ref{eq:NormTune_quadratic_scored} into Eq.~\ref{eq:NormTune_general_mappingd}, we obtain the final bucket-wise learning rate:
\begin{equation}
    \eta_k
    =
    \eta_{\min}
    +
    (\eta_{\max}-\eta_{\min})
    \left(1-q_k^{2}\right).
    \label{eq:bucket_lrd}
\end{equation}

All layers in the same bucket share the same learning rate:
\begin{equation}
    \eta_l=\eta_k,
    \qquad
    \text{if layer } l \text{ belongs to bucket } k .
    \label{eq:NormTune_layer_assignmentd}
\end{equation}
As shown in Figure~\ref{fig:peft_our_norm}, our method consistently increases the update norms of insufficiently adapted layers, aligning them more closely with full fine-tuning and improving model performance.

\section{Experimental Results}
\paragraph{Datasets.}
For training, we use the S1K-1.1 dataset \cite{Muennighoff2025}, the Z1-Code-Reasoning107K dataset \cite{Yu2025Z1}, the Data-repetition dataset \cite{kopiczko2026data}, the GovReport dataset for long-context understanding\cite{huang2021efficient}, and the Commonsense Reasoning dataset for short reasoning.
We further evaluate different models on several challenging tasks. The reasoning-intensive tasks consist of AIME 2024 \cite{aime2024}, MATH 500 \cite{hendrycks2021measuring}, GPQA \cite{rein2024gpqa}, and AIME 2025 \cite{aime2025}. The code-scientific reasoning tasks consist of MATH 500, GPQA, LiveCodeBench \cite{jain2024livecodebench}, and BigCodeBench-Hard \cite{zhuo2024bigcodebench}.

\paragraph{Compared Methods.}

The baseline methods are broadly divided into two categories: reinforcement learning (RL) and supervised fine-tuning (SFT). In the RL category, we compare against OpenAI o1 \cite{OpenAI2024}, o3 \cite{openai2025o3mini}, DeepSeek-R1 \cite{guo2025deepseek}, and QwQ-32B \cite{qwen2025qwq32b}. In the SFT category, we compare against S1(Full-FT) \cite{Muennighoff2025}, Z1-7B(Full-FT) \cite{Yu2025Z1}, Qwen3-4B(Full-FT) \cite{kopiczko2026data} and  OLMo-3-7B(Full-FT) \cite{kopiczko2026data}.
In addition, we compare against several representative PEFT methods, including DoRA~\cite{liu2024dora}, AdaLoRA~\cite{zhang2023adalora}, LoRA+~\cite{hayou2024lora+}, rsLoRA~\cite{kalajdzievski2023rank}, LoRA  \cite{hu2021lora}, Bone\cite{kang2024bone} and MoRA\cite{jiang2024mora}. We further incorporate TempBalance~\cite{zhou2023temperature} into the PEFT baselines for comparison.

\paragraph{Evaluation Metrics.}


For the S1K-1.1 and Z1-Code-Reasoning107K settings, we evaluate each locally evaluated checkpoint using greedy decoding with temperature \(T=0\), generating one completion per problem in each evaluation run. We repeat evaluation on the same checkpoint and the same benchmark questions for \(R=4\) runs on MATH 500 and GPQA, and \(R=16\) runs on AIME 2024, AIME 2025, LiveCodeBench, and BigCodeBench-Hard, where applicable. Tables \ref{main_table} and \ref{Z1_table} report the mean accuracy across these evaluation runs. These repetitions are evaluation runs of a fixed checkpoint, rather than independent training runs or stochastic decoding samples. The Avg. column is the unweighted mean of the four benchmark accuracies. For Qwen3-4B and OLMo-3-7B, we retain the sampling-based Avg@k and Pass@k metrics under the decoding configurations specified in Table \ref{tab:evaluation_settings}. Results taken from prior publications retain their original evaluation protocols and are explicitly distinguished from locally evaluated results. Details of the uncertainty estimation are provided in Table \ref{tab:evaluation_settings}.
For the environment and settings also provided in Appendix \ref{Environment and Settings}.

\begin{table*}[t]
  \caption{Accuracy (\%) of different test-time scaling methods on reasoning-intensive tasks with Qwen2.5-32B-Instruct. $^{*}$ denotes results reported in \cite{Muennighoff2025}. The results of T1 (Qwen2.5-32B) and SSFT-32B-GFPO are sourced from \cite{hou2025t1} and \cite{jia2025training}.}
  \label{main_table}
  \begin{center}
    \begin{small}
      \begin{sc}
      \begin{adjustbox}{scale=0.7}
        \begin{tabular}{llccccccc}
\toprule \multicolumn{1}{c}{\multirow[c]{2}{*}{Dataset}} & \multicolumn{1}{c}{\multirow[c]{2}{*}{Method}} & \multirow[c]{2}{*}{Rank} & AIME 2024 & MATH 500 & GPQA & AIME 2025 & \multirow[c]{2}{*}{Avg.} \\ & & & Avg@16 & Avg@4 & Avg@4 & Avg@16 & \\ \midrule
          

          \multirow{35}{*}{S1K-1.1}
          &o3-mini-low$^{*}$ &-& 56.3 &95.8 &70.6& 42.1& 66.2 \\
          &QwQ-32B$^{*}$ &- & 46.7& 90.6& 54.5& 32.7 & 56.13 \\
          & T1 (Qwen2.5-32B)& -& 50.6&92.4& 56.1&-&- \\
          & SSFT-32B-GFPO&-& 64.22& 89.90& 62.48&   58.80& 68.85\\
          &r1$^{*}$ &- &   79.8 &97.3& 71.5 &70.0 & 79.65 \\
          \cmidrule(lr){2-8}
          &S1 (Full-FT)$^{*}$          & - & 56.7 & 95.4 & 63.6&50 &66.43   \\
          &r1-distill-Llama-70B$^{*}$  &-&  57.1& 94.5& 65.2 & 56.3 & 68.28 \\
          & r1-distill-Qwen-14B$^{*}$ &-&  61.7 &93.9 &59.1& 48.0 & 65.68 \\          
          &  r1-distill-Qwen-32B$^{*}$ &-&  58.3  &94.3& 62.1&  49.6   & 66.08 \\

         & AdaLoRA        & 8 &6.67	&67.4	&27.27	&6.67	&27.00\\
          & DoRA        & 8 &26.67	&78.8	&47.98	&6.67	&40.03\\
           & RSLoRA        & 8 & 16.67	&74.6	&50.51	&0	&35.45\\
           & LoRA+        & 8 & 33.33	&88.8	&51.52	&33.33	&51.75\\

        & LoRA        & 8 &10&	78.8&	47.47&	13.33&	37.4\\
        & TempBalance + LoRA       & 8 &20 &	80.4	&51.01	&13.33	&41.19 \\
          & \textbf{NormTune+LoRA(Ours)  }       &8  &   53.33 &	92.6 &	60.61 &	40 &	61.64  \\
          & Bone        & 8 &26.67&	80.4&	52.53&	40 &	49.9\\
          & TempBalance + Bone       & 8 &36.67	&89.4&	55.05&	40&	55.28 \\
          & \textbf{NormTune+ Bone (Ours) }       & 8 & 50 &	92.8&	62.12&	40&	61.23 \\
           & MoRA        & 8 &26.67	&80.4&	51.01	&20&	44.52 \\
           & TempBalance + MoRA       & 8 & 23.33	 &89.2	&51.01	&26.67	&47.55 \\
           & \textbf{NormTune+MoRA (Ours) }     & 8 & 40 &	91.6	 &64.65 &	46.67	 &60.73  \\

           & AdaLoRA        & 256 &13.33  & 	67.4	  & 37.37  & 	6.67  & 	31.19\\
           & DoRA        & 256 &23.33	&88.6	&53.03	&33.33	&49.57\\
           & RSLoRA        & 256& 43.3	&90.2	&60.1	&33.3	&56.73 \\
         & LoRA+        & 256 & 50&	90.6&	58.59&	33.33&	58.13\\

          &LoRA        & 256 &36.67&	92.6	&58.08	&33.33&	55.16\\
          & TempBalance + LoRA       & 256   & 40  & 90.2&	54.4	&40	&56.15\\
          &\textbf{NormTune+LoRA (Ours)}        & 256 & 63.33&	93.6&	64.14&	53.33&	\textbf{68.6}\\
          &Bone        & 256 &43.33	&93.4	&57.58	&33.3&	56.90\\
            & TempBalance + Bone       & 256 & 50	&90.8&	60.61	&40	&60.35 \\
          &\textbf{NormTune+Bone (Ours) }     & 256 &
          56.67&	94.8&	64.65&53.33 &\textbf{67.36}\\
          &MoRA      & 256 &50 &	93.2 &	62.12 &	33.3 &	59.66  \\
          & TempBalance + MoRA       & 256 & 56.67	&93.2	&62.63&	33.33&	61.46 \\
          &\textbf{NormTune+MoRA  (Ours) }    & 256 &60&	94.6&	63.13&	60	&\textbf{69.43}\\
          

          

          \bottomrule
        \end{tabular}
        \end{adjustbox}
        \vspace{1.5ex}
      \end{sc}
    \end{small}
  \end{center}
  \vskip -0.1in
\end{table*}

\begin{table*}[t]
  \caption{Accuracy (\%) of  different test-time scaling methods on code-scientific reasoning tasks with Z1.  $^{*}$ denotes results reported in \cite{Yu2025Z1}.}
  \label{Z1_table}
  \begin{center}
    \begin{small}
      \begin{sc}
          \begin{adjustbox}{scale=0.7}
        \begin{tabular}{llccccccc}
          \toprule
Dataset & Method & Rank
& \shortstack{MATH 500\\AVG@4}
& \shortstack{GPQA\\AVG@4}
& \shortstack{LiveCodeBench\\AVG@16}
& \shortstack{BigCodeBench-Hard\\AVG@16}
& Avg. \\
          
          \midrule
          \multirow{20}{*}{\rotatebox[origin=c]{90}{Z1-Code-Reasoning107K}}
          
           & o1-preview$^{*}$ &-& 85.5& 73.3& 43.2& 23.0 &56.3      \\
           
           &o1-mini$^{*}$      & - & 90.0  & 60.0 &  53.7 &  27.7 &  57.9\\
           
           &Deepseek-R1$^{*}$     & -  &97.3 &71.5& 77.9& 29.7 &67.6\\

&QwQ-32B-Preview$^{*}$   & - & 90.6 &60.0& 59.9 &25.0& 58.9\\
           
           & Deepseek-V3$^{*}$     & -  & 90.2 & 59.1&  56.3&  27.7&  58.3\\
& GPT-4o-0513$^{*}$     & -  & 75.8&  46.5 & 43.4&  25.0&  47.7\\

\cmidrule(lr){2-8}
    & Z1-7B (Full-FT)$^{*}$     & -   &76.4& 47.5 &35.3 &22.3 & \underline{45.4}       \\
    &R1-Distill-Qwen-7B$^{*}$     & -  &83.3& 49.1& 40.5& 3.4 &44.1\\
    & s1.1-7B$^{*}$     & -  & 79.2 & 31.8&  15.2 & 4.7&  31.7\\
    & OpenThinker-7B$^{*}$     & -  & 83.0 & 42.4 & 25.3&  17.6&  42.1\\

        & AdaLoRA        & 256& 72	&51.5&	29.4&	12.8 &	41.4\\
          & DoRA         & 256& 73.2	&49.5	&30.2	&18.2&	42.8\\
           & RSLoRA        & 256& 72.4	&	46.5&		29.7&		13.5	&	40.5\\
           & LoRA+        & 256&72	&49.5	&30.9	&17.6	&42.5\\
         
    & LoRA & 256 &  73.4 & 50 &  30.3 &   10.8 & 41.1  \\
    & TempBalance + LoRA       & 256 & 72.4	&53.5&	32.3&	16.2	&43.6 \\
          & \textbf{NormTune+LoRA(Ours)}  & 256 &   75.2  &	50 &36.3   &22.3 & \textbf{46.0}   \\
          
          &Bone & 256 & 75.4 & 46.5 & 32.3 & 17.6 &  43.0  \\
            & TempBalance + Bone       & 256   &   75.2	 & 49.5	 & 33.5	 & 16.9	 & 43.8\\
          &  \textbf{NormTune+Bone (Ours)} & 256 &   76.8 &	47.5    & 34.2      &    22.7   &   45.3       \\
       &   MoRA      & 256 & 71.6&	44.4&	30.9&	17.6& 41.1\\
    & TempBalance + MoRA   & 256   & 76.6	&46	&32.6	&18.2	&43.4\\
          &\textbf{ NormTune+MoRA(Ours)  }   & 256 & 76.6	&50	&35.4&	19.6	& \textbf{45.4}  \\
          \bottomrule
        \end{tabular}
        \end{adjustbox}
      \end{sc}
    \end{small}
  \end{center}
  \vskip -0.1in
\end{table*}

\begin{table*}[t]
\centering
\caption{Performance of Full-FT, PEFT, and NormTune-enhanced PEFT methods on OLMo-3-7B across GPQA, AIME'24, and AIME'25. $^{*}$ denotes results reported in \cite{kopiczko2026data}.}
\label{datarepet_peft}
\resizebox{0.9\textwidth}{!}{
\begin{tabular}{ l c |cc|cc|cc|c}
\toprule
 Model & Method & \multicolumn{2}{c|}{GPQA} & \multicolumn{2}{c|}{AIME'24} & \multicolumn{2}{c|}{AIME'25} & Average \\
&&  Avg@4 & Pass@4 & Avg@16 & Pass@16 & Avg@16 & Pass@16 & \\
\midrule
\multirow{14}{*}{\rotatebox[origin=c]{90}{OLMo-3-7B}}
& Full-FT$^{*}$ & 34.0 & 62.1 & 42.3 & 80.0 & 39.2 & 63.3 & 53.5\\
& AdaLoRA  &    42.6 &	66.2 &	22.1 &	43.3	 &22.7 &	50 &	41.2 \\
          & DoRA         &30.6	&57.5	&33.3&	73.3	&30.8&	56.7	&47.0 \\
           & RSLoRA        & 	42.6& 	70.2	& 22.7	& 56.6	& 21.7& 	53.3	& 44.5\\
           & LoRA+        &40.9	&65.2&	21.3&	53.3&	16.5&	46.7	&40.7 \\
& LoRA(256)& 31.9&	62.1&	26.7&	63.3&	24&	43.3&	41.9 \\
 & TempBalance + LoRA       & 40.9	  & 65.2  & 	21.3  & 	53.3	&20.2&	56.7	&42.9\\
&\textbf{NormTune+LoRA(256)  } & 38.8&	69.2&	42.3	&80&	35.4	&60&	\textbf{54.3}\\
& Bone(256)& 35.9&	63.6&	33.3&	73.3&	31&	56.7&	49.0 \\
& TempBalance + Bone       &30.6	 &57.5	 &35.4	 &66.7	 &38.5	 &67.2 &	49.3 \\
&\textbf{NormTune+Bone(256)} & 38.1& 	60& 	45.2& 	76.7& 	37.1& 	64.6& 	\textbf{53.6} \\
& MoRA(256)& 37.6	& 65.7& 	37.5& 	76.7& 	29.4& 	50& 	49.5 \\
& TempBalance + MoRA       &32.7	&60	&42.3	&76.7	&35.8	&63.3	&51.8 \\
& \textbf{NormTune+MoRA(256) }  & 39.9 & 	68.7 & 	41 & 	76.7 & 	35.8 & 	63.3 & 	\textbf{54.2}\\

\bottomrule
\end{tabular}
}
\end{table*}

\subsection{Main Comparison Results}
In this section, we present the main comparison results by applying NormTune to representative PEFT methods, including LoRA, Bone, and MoRA, and evaluating their effectiveness for test-time scaling. 
As shown in Table~\ref{main_table}, rank-8 PEFT methods underperform on reasoning-intensive tasks. Incorporating NormTune substantially improves their performance, yielding average gains of 24.24\%, 11.33\%, and 16.21\% for LoRA, Bone, and MoRA, respectively. Rank-8 NormTune-enhanced PEFT surpasses 90\% of full fine-tuning performance and consistently outperform AdaLoRA, LoRA+, rsLoRA, DoRA and TempBalance+PEFT under the same rank-8 setting.
When increasing the rank to 256, NormTune still brings consistent improvements to PEFT methods, with average gains of 13.44\%, 10.46\%, and 9.77\% for LoRA, Bone, and MoRA, respectively. Notably, all three NormTune-enhanced variants surpass S1 (Full-FT) by 2.17\%, 0.93\%, and 3.00\%, respectively. 
Beyond supervised fine-tuning, we further compare NormTune with state-of-the-art reinforcement-learning-based reasoning models.
Using the same Qwen2.5-32B-Instruct backbone, all NormTune-enhanced variants outperform T1 (Qwen2.5-32B), and NormTune+MoRA further surpasses SSFT-32B-GFPO by 0.58\%. These results show that NormTune substantially improves model performance with lower resource consumption.

For Z1-7B based one Qwen-2.5-Coder-7B-Instruct, we further evaluate code-related reasoning performance, as shown in Table~\ref{Z1_table}. 
Under supervised fine-tuning with a similar model size, NormTune+LoRA, NormTune+Bone, and NormTune+MoRA all achieve performance comparable to the state-of-the-art full fine-tuning baseline Z1-7B (Full-FT). Notably, NormTune+LoRA further surpasses Z1-7B by over 0.6\%. For RL-based training, NormTune+LoRA also achieves a performance level comparable to GPT-4o-0513, with an average score of 46.0 versus 47.7. This demonstrates the effectiveness of NormTune across various testing datasets.

For OLMo-3-7B, as shown in Table~\ref{datarepet_peft}, all NormTune-enhanced methods, including NormTune+LoRA, NormTune+Bone, and NormTune+MoRA, outperform Full-FT. 
These results show that NormTune remains effective even for smaller models. Qwen3-4B experiments are presented in the Appendix in Table \ref{tab:budget_results}.

Across scales from 4B to 32B and across mathematical reasoning, code reasoning, and scientific question-answering benchmarks, NormTune consistently improves different PEFT methods. This demonstrates that NormTune is a general and scalable solution for improving test-time scaling with strong performance.


\subsection{Ablation Study}

\begin{table}[t]
  \centering

  \begin{minipage}[c]{0.48\linewidth}
  
    \caption{Accuracy (\%) of PEFT methods under various adjustment styles and ranks on S1. }
    \label{albadecay_table}
    \centering
    \setlength{\tabcolsep}{2pt}
\begin{adjustbox}{max width=\linewidth, scale=0.7}

      \begin{tabular}{lccr}
        \toprule
        METHOD & \shortstack{ADJUST.\\STYLE} & RANK & Avg.\\
        \midrule
        Full-FT & - & - & 66.43\\
        Full-FT & Stage & - & 46.43\\
        \midrule
        LoRA & - & 256 & 55.16\\
        LoRA & Linear & 256 & 60.76\\
        LoRA & Stage & 256 & 63.2\\
        LoRA & Exponential & 256 & 62.79\\
        LoRA & NormTune & 256 & \textbf{68.6}\\
        Bone & - & 256 & 56.90\\
        Bone & Linear & 256 & 59.79\\
        Bone & Stage & 256 & 63.28\\
        Bone & Exponential & 256 & 60.17\\
        Bone & NormTune & 256 & \textbf{67.36}\\
        MoRA & - & 256 & 59.66\\
        MoRA & Linear & 256 & 61.79\\
        MoRA & Stage & 256 & 60.78\\
        MoRA & Exponential & 256 & 63.03\\
        MoRA & NormTune & 256 & \textbf{69.43}\\
        \bottomrule
      \end{tabular}
    \end{adjustbox}
  \end{minipage}
  \hfill
  \begin{minipage}[c]{0.48\linewidth}
    \centering
    \setlength{\tabcolsep}{2pt}
    \caption{Accuracy (\%) of full fine-tuning and rank-256 PEFT methods under two learning-rate settings on reasoning-intensive tasks. See Table~\ref{differate-tabledddd} for detailed results.}
    \label{differate-table}
    \begin{adjustbox}{max width=\linewidth, scale=0.9}
      \begin{tabular}{lccr}
        \toprule
        Method & Learning rate & Rank & Avg.\\
        \midrule
        S1 (Full-FT) & 1e-5 & - & 66.43\\
        S1 (Full-FT) & 1e-4 & - & 4.9\\
        LoRA & 1e-5 & 256 & 55.16\\
        LoRA & 1e-4 & 256 & 49.7\\
        Bone & 1e-5 & 256 & 56.90\\
        Bone & 1e-4 & 256 & 55.65\\
        MoRA & 1e-5 & 256 & 59.66\\
        MoRA & 1e-4 & 256 & 49.47\\
        \bottomrule
      \end{tabular}
    \end{adjustbox}
  \end{minipage}
\end{table}

\paragraph{Learning Rate.}
We fine-tune Qwen2.5-32B-Instruct on the S1K-1.1 reasoning dataset with different rank-256 PEFT methods under two learning-rate settings. As shown in Table~\ref{differate-table}, increasing the learning rate from $1\times10^{-5}$ to $1\times10^{-4}$ consistently hurts performance: the average accuracy decreases from 55.16\% to 49.70\% for LoRA, from 56.90\% to 55.65\% for Bone, and from 59.66\% to 49.47\% for MoRA. Full fine-tuning also collapses under the larger learning rate, with the average score dropping from 66.43\% to 4.90\%. These results suggest that performance gains cannot be obtained by uniformly increasing the learning rate, motivating NormTune to adaptively allocate larger update scales only to under-adapted layers.



\begin{figure}[t]
    \centering
    \begin{subfigure}[t]{0.48\linewidth}
        \centering
        \includegraphics[width=0.9\linewidth]{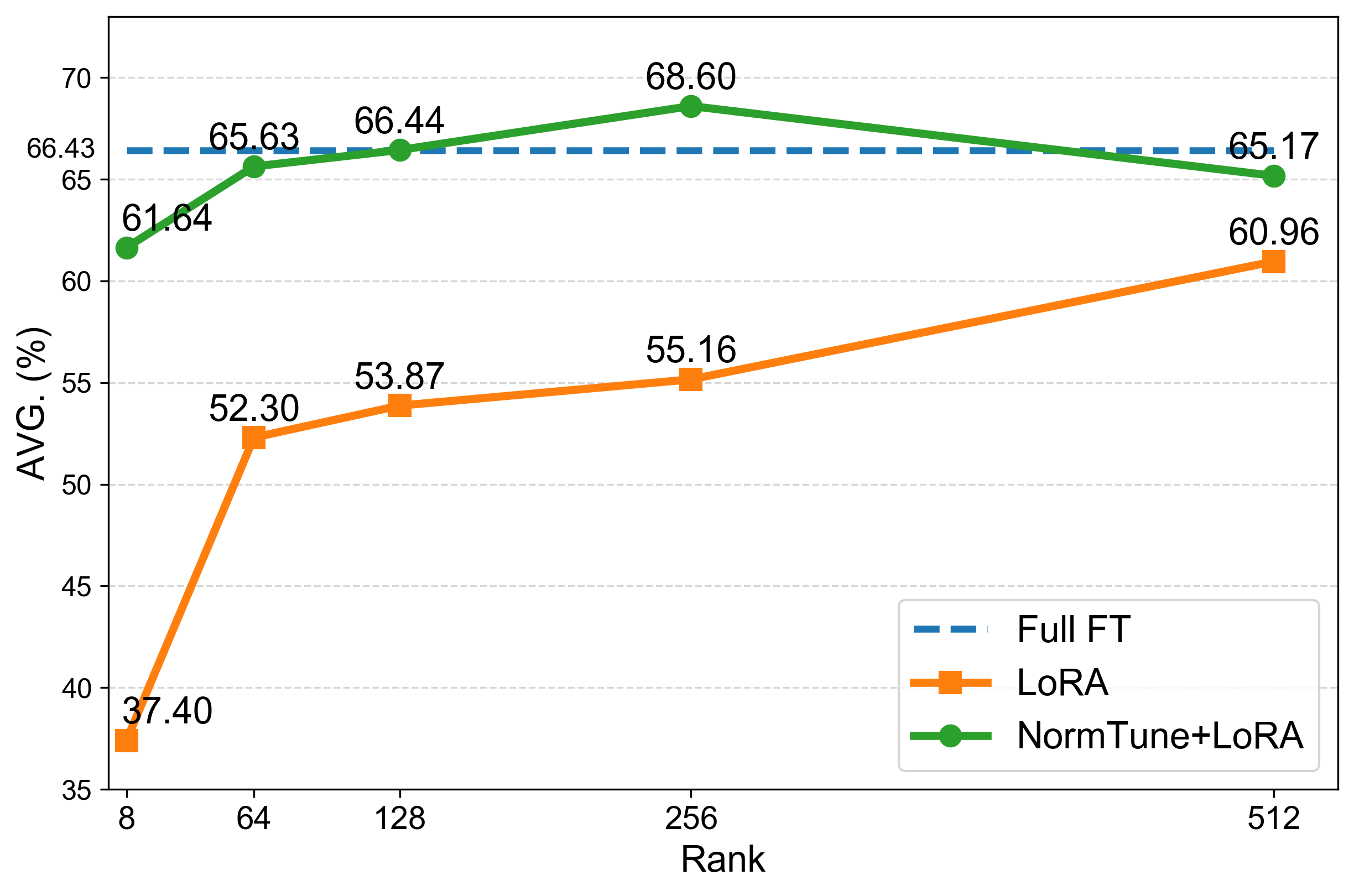}
        \caption{Average performance across reasoning-intensive tasks under different rank settings.}
        \label{fig:rank_analysis}
    \end{subfigure}
    \hfill
    \begin{subfigure}[t]{0.48\linewidth}
        \centering
        \includegraphics[width=0.9\linewidth]{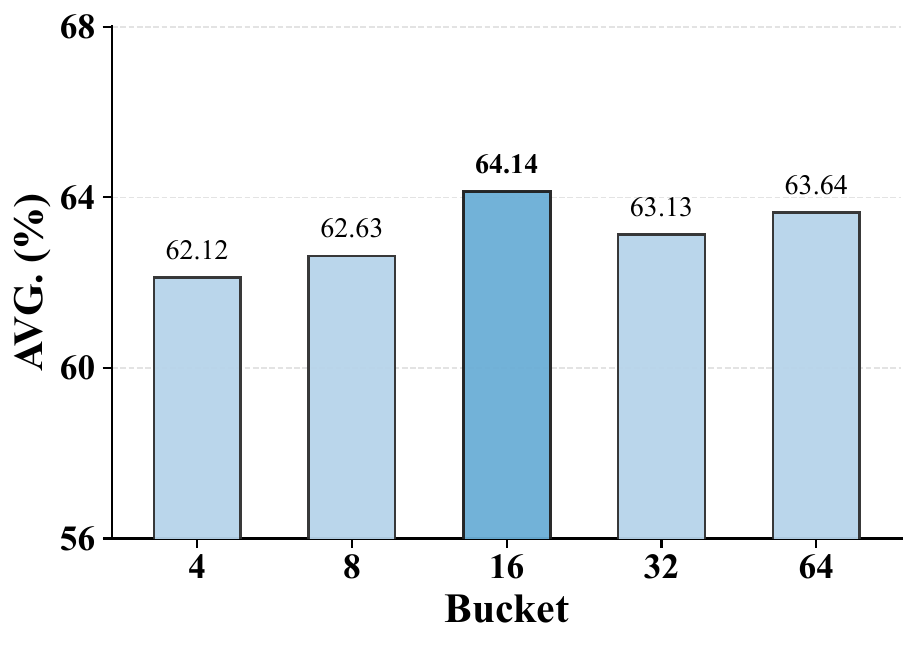}
        \caption{Effect of the NormTune bucket number $n$ on GPQA.}
        \label{fig:aime_n_ablation}
    \end{subfigure}
    \caption{Results after training on S1K-1.1.}
    \label{fig:rank_bucket_analysis}
\end{figure}

\paragraph{Rank Sensitivity.}
As shown in Figure~\ref{fig:rank_analysis}, increasing the rank generally improves both LoRA and NormTune+LoRA, while NormTune+LoRA consistently outperforms LoRA across all ranks. The gains are particularly pronounced in low-rank regimes: at rank 8, NormTune+LoRA improves the average score from 37.40\% to 61.64\%. NormTune+LoRA achieves its best average performance at rank 256, reaching 68.6\% and slightly exceeding the Full-FT baseline of 66.43\%. Further increasing the rank to 512 reduces the score to 65.17\%, indicating that larger ranks do not necessarily yield better reasoning-intensive performance. These results show that NormTune effectively improves LoRA across different rank budgets, especially when the adaptation capacity is highly constrained.

\paragraph{Adjustment Style.}
Based on the norm distribution of PEFT methods on reasoning-intensive tasks, and inspired by \cite{howard2018universal}, we further propose three naive layer-wise learning-rate adjustment strategies. The first one uses a linear decay across layers:
$\eta_i = \eta_{\mathrm{max}} + 
(\eta_{\mathrm{min}} - \eta_{\mathrm{max}})
\cdot \frac{i}{N-1},
$
where $i$ is the layer index and $N$ is the total number of layers. 
The second strategy, termed stage decay, assigns $\eta_{\mathrm{max}}$ to layers below layer 45 and $\eta_{\mathrm{min}}$ to layer 45 and above, since the former exhibit under-adaptation in layer-wise norms, as shown in Figure~\ref{fig:vs_total_4b}. The last one, we further evaluate exponential-decay strategy, $e^{\frac{\max_i(norm_i)}{norm_i}-1}$. 
As shown in Table~\ref{albadecay_table}, both linear, stage and  exponential-decay strategies improve the performance of all PEFT methods over their original versions. However, their gains are consistently smaller than those achieved by NormTune. For example, NormTune improves the average scores to 68.6\%, 67.36\%, and 69.43\% for LoRA, Bone, and MoRA, respectively, outperforming both linear, stage and  exponential-decay strategies. See Table~\ref{albadecay_tabledddd} for detailed results. These results indicate that simply imposing a predefined learning-rate decay is beneficial, but the adaptive design of NormTune is more effective for reasoning-intensive adaptation. Moreover, full fine-tuning generally provides sufficient adaptation across layers, making it less suitable for our stage decay strategy.





\subsection{Hyperparameter Sensitivity Analysis}

We analyze the sensitivity of NormTune to the bucket number $n$, which determines the granularity of layer-wise learning-rate allocation. As shown in Figure~\ref{fig:aime_n_ablation}, NormTune is robust to a wide range of $n$. Across $n \in \{4,8,16,32,64\}$, the GPQA accuracy only changes from 62.12\% to 64.14\%, showing limited sensitivity to this hyperparameter. The best result is obtained at $n=16$, where NormTune achieves 64.14\%. This indicates that a moderate bucket number provides sufficient layer-wise differentiation without introducing unnecessary fragmentation.




\section{Conclusion}
In this work, we studied parameter-efficient fine-tuning for reasoning-intensive tasks with test-time scaling. 
Through three key observations,  we propose Norm-Guided Parameter-Efficient Fine-Tuning (NormTune), which adaptively enhances under-trained layers and thereby substantially improves performance on reasoning-intensive tasks.
Extensive experiments on three benchmarks demonstrate that NormTune consistently improves different PEFT methods and can achieve performance comparable to full fine-tuning with substantially lower resource consumption.


\subsection*{AI use statement}
Generative AI was used solely to assist with the organization and formatting
of early manuscript drafts. It was not used for method conception, experimental implementation,
or the substantive writing of the paper.

\subsection*{Ethics statement}
This work investigates Norm-Guided Parameter-Efficient Fine-Tuning for reasoning-intensive tasks. We do not involve human subjects or introduce new datasets. The primary foreseeable impact is improved performance of PEFT methods on reasoning-intensive tasks.

\subsection*{Reproducibility statement}
An anonymized supplementary package will include the exact coordinate,
calibration, reconstruction, and terminal-evaluation configurations; fixed
data splits and random seeds; per-cell manifests; and fresh-process evaluation
scripts.

\bibliography{iclr2027_conference}
\bibliographystyle{iclr2027_conference}

\appendix
\section{Appendix}
\label{sec:appendix}

This appendix provides related work, additional results and implementation details to complement the main paper. It is organized as follows. 
Appendix~\ref{Related Work} introduce the related work. Appendix~\ref{Experiments} presents extended experimental results, including reasoning-intensive evaluations on Qwen3-4B and additional analyses across ranks and task types. Appendix~\ref{reasoning-intensive Experiments} further evaluates NormTune on GPQA, AIME'24, and AIME'25, while Appendix~\ref{Other Experiments} reports detailed results corresponding to the rank analysis and task comparison in the main paper. Appendix~\ref{Environment and Settings} describes the additional benchmarks, experimental settings, and hardware configurations used in our experiments. 
Finally, Appendix~\ref{Prompts for Evaluation} describes the Prompts for Evaluation.

\subsection{Related Work}
\label{Related Work}

\subsubsection{Test-Time Scaling for Reasoning}

Test-time scaling  \citep{OpenAI2024,DeepSeek2025,openai2025o3mini,bsharat2025prompting,xu2025double,kopiczko2026data} has recently emerged as an effective way to improve reasoning performance by allocating more computation at inference time. Existing methods can be broadly grouped into two categories: reinforcement learning (RL)~\citep{openai2025o3mini,OpenAI2024,DeepSeek2025} and supervised fine-tuning (SFT)\citep{bsharat2025prompting,xu2025double,kopiczko2026data}. For reinforcement learning (RL), 
despite their strong performance, these methods often depend on expensive post-training pipelines, are challenging to reproduce, or involve additional inference-time search components.

To address these limitations, recent studies have explored simpler and more reproducible forms of test-time scaling based on supervised fine-tuning \citep{xu2025double,an2025long,kopiczko2026data}.
Although these methods are generally easier to reproduce and adopt than reinforcement-learning-based approaches, training them still requires substantial computational resources, especially for large models, and often depends on access to high-memory GPUs.

\subsubsection{Parameter-Efficient Fine-Tuning}

Parameter-efficient fine-tuning (PEFT)\citep{houlsby2019parameter,li2021prefix,liu2024dora,feng2025omoe,yao2024layer} is introduced to adapt pretrained models without updating all backbone parameters, thereby substantially reducing GPU memory consumption.
Early PEFT methods\citep{houlsby2019parameter,li2021prefix} add small task-specific modules or prompts while keeping the main model frozen. 
However, these methods often introduce additional inference latency and increase the computational cost of generation.
To address this issue, low-rank adaptation\citep{zhang2023adalora,kalajdzievski2023rank,hayou2024lora+, hu2021lora,liu2024dora,jiang2024mora}, which can be merged into the backbone model during inference, has become the most widely used PEFT approach for large language models. 
Although these methods greatly improve memory efficiency during fine-tuning, they often underperform full fine-tuning, especially under low-rank settings. In this paper, we propose Norm-Guided Parameter-Efficient Fine-Tuning (NormTune), which adaptively enhances under-trained layers and thereby substantially improves performance on reasoning-intensive tasks.

\subsection{Experiments}
\label{Experiments}

\subsubsection{Reasoning-Intensive Experiments}
\label{reasoning-intensive Experiments}

We further evaluate NormTune on Qwen3-4B across GPQA, AIME'24, and AIME'25 benchmarks. As shown in Table~\ref{tab:budget_results}, NormTune consistently improves different PEFT backbones across model families and reasoning benchmarks. On Qwen3-4B, NormTune+LoRA improves the average score from 37.7\% to 44.2\%, surpassing Full-FT by 6\%, while NormTune+MoRA improves MoRA from 33.9\% to 42.9\%. NormTune+Bone also achieves consistent gains and reaches performance comparable to Full-FT. However, TempBalance+LoRA, TempBalance+Bone, and TempBalance+MoRA reduced performance by 0.7\%, 11\%, and 17.3\% compared with LoRA, Bone, and MoRA, respectively.



\begin{table*}[t]
  \caption{Accuracy (\%) of LoRA and NormTune+LoRA with different ranks on reasoning-intensive tasks}
  \label{albations_table}
  \begin{center}
    \begin{small}
      \begin{sc}
      \begin{adjustbox}{max width=\textwidth}
        \begin{tabular}{llcccccc}

\toprule \multicolumn{1}{c}{\multirow[c]{2}{*}{Model}} & \multicolumn{1}{c}{\multirow[c]{2}{*}{Method}} & \multirow[c]{2}{*}{Rank} & AIME 2024 & MATH 500 & GPQA & AIME 2025 & \multirow[c]{2}{*}{Avg.} \\ & & & Avg@16 & Avg@4 & Avg@4 & Avg@16 & \\ \midrule

          \multirow{10}{*}{\parbox{2.6cm}{\centering Qwen2.5-32B\\-Instruct}}
          
          &Full-FT           & - & 56.7 & 95.4 & 63.6&50 &66.43   \\
          &LoRA         & 8 &10&	78.8&	47.47&	13.33&	37.4\\
          &NormTune+LoRA        &8 &   53.33 &	92.6 &	60.61 &	40 &	61.64\\
          &LoRA      & 64 & 30& 90.8&55.05 &33.33 & 52.3 \\
          &NormTune+LoRA       &64 & 66.67& 	90.6& 	58.59& 	46.67& 	65.63  \\
        &    LoRA      & 128 & 26.67 &	90.2&	58.59	&40	&53.87 \\
         &    LoRA +NormTune        &128  & 60	&91.8&	60.61&	53.33&	66.44\\
          &LoRA       & 256 &36.67&	92.6	&58.08	&33.3&	55.16\\
          &LoRA +NormTune        & 256 & 63.33&	93.6&	64.14&	53.33&	\textbf{68.6}\\
          &LoRA       & 512 & 40&	91.4&	59.09&	53.33&	60.96\\
          &LoRA +NormTune      & 512  &56.67 &	94.2&	63.13&	46.67	& 65.17\\
          
          \bottomrule
        \end{tabular}
        \end{adjustbox}
      \end{sc}
    \end{small}
  \end{center}
  \vskip -0.1in
\end{table*}

\begin{table*}[t]
\centering
\caption{Performance of Full-FT, PEFT, and NormTune-enhanced PEFT methods on Qwen3-4B across GPQA, AIME'24, and AIME'25. $^{*}$ denotes results reported in \cite{kopiczko2026data}.}
\label{tab:budget_results}
\resizebox{0.9\textwidth}{!}{
\begin{tabular}{ l c |cc|cc|cc|c}
\toprule
 Model & Method & \multicolumn{2}{c|}{GPQA} & \multicolumn{2}{c|}{AIME'24} & \multicolumn{2}{c|}{AIME'25} & Average \\
&&  Avg@4 & Pass@4 & Avg@16 & Pass@16 & Avg@16 & Pass@16 & \\
\midrule
\multirow{14}{*}{\rotatebox[origin=c]{90}{Qwen3-4B}}
& Full-FT$^{*}$   & 39.3 &68.7 &19.2 & 43.3 & 18.8 & 40.0 & 38.2 \\
& AdaLoRA        & 21.8	 &54 &	7.1 &	30 &	9.8	 &26.7	 &24.9\\
          & DoRA         &38.1&	72.2&	15	&33.3	&16.3	&43.3&	36.4 \\
           & RSLoRA        & 21.3	&51.5&	6.9&	26.7&	5.8	&20	&22.0	\\
           & LoRA+        & 27	&62.1	&12.5&	36.7	&8.1	&30	&29.4\\
& LoRA(256)&  39&	66.2&	15&	40&	19.4&	46.7&	37.7\\
 & TempBalance + LoRA       & 38.8 &	70.7 &	14.2 &	40	 &14.8	 &43.3	 &37\\
& \textbf{NormTune+LoRA(256)  } & 42.6&	70.2&	{22.7}&	{56.6}&	{20.0}&	{53.3}&\textbf{44.2}	
\\
& Bone(256)& 35.1& 	63.6& 	15& 	33.3& 	14.17& 	40	& 33.5\\
& TempBalance + Bone       & 27.3	&59.6	&4.6	&20&	3.6	&20	&22.5\\
& \textbf{NormTune+Bone(256)}  & {42.3} & 	67.7 & 	16.5 	 & 36.7	 & {21.9}	 & 50 & \textbf{39.2} \\
& MoRA(256)& 38.3&	69.7&	15.2&	30	&10&	40&	33.9 \\
& TempBalance + MoRA       &23.4&	58.6&	0.4	&6.7	&0.6	&10	&16.6 \\
& \textbf{NormTune+MoRA(256) }  & 41.3 &	{71.7} &	 {20.8} &	{46.7} &	20.2 &	{56.7} &	\textbf{42.9}   \\
\bottomrule
\end{tabular}
}
\end{table*}

\begin{table*}[t]
  \caption{Accuracy (\%) of LoRA with different ranks on reasoning-intensive tasks.}
  \label{alb_table_rankdddd}
  \begin{center}
    \begin{small}
      \begin{sc}
      \begin{adjustbox}{max width=\textwidth}
        \begin{tabular}{llcccccc}

          \toprule \multicolumn{1}{c}{\multirow[c]{2}{*}{Model}} & \multicolumn{1}{c}{\multirow[c]{2}{*}{Method}} & \multirow[c]{2}{*}{Rank} & AIME 2024 & MATH 500 & GPQA & AIME 2025 & \multirow[c]{2}{*}{Avg.} \\ & & & Avg@16 & Avg@4 & Avg@4 & Avg@16 & \\ \midrule

          \multirow{6}{*}{ \parbox{2.6cm}{\centering Qwen2.5-32B\\-Instruct}}
          
          &Full-FT           & - & 56.7 & 95.4 & 63.6&50 &66.43   \\
          &LoRA         & 8 &10&	78.8&	47.47&	13.33&	37.4\\
          &LoRA      & 64 & 30& 90.8&55.05 &33.33 & 52.3 \\
        &    LoRA      & 128 & 26.67 &	90.2&	58.59	&40	&53.87 \\
          &LoRA       & 256 &36.67&	92.6	&58.08	&33.33&	55.16\\
          &LoRA       & 512 & 40&	91.4&	59.09&	53.33 & 60.96 \\
          
          \bottomrule
        \end{tabular}
        \end{adjustbox}
      \end{sc}
    \end{small}
  \end{center}
  \vskip -0.1in
\end{table*}

\begin{table*}[t]
  \caption{Accuracy (\%) of full fine-tuning and rank-256 PEFT methods under two learning-rate settings on reasoning-intensive tasks.}
  \label{differate-tabledddd}
  \begin{center}
    \begin{small}
      \begin{sc}
      \begin{adjustbox}{max width=\textwidth}
        \begin{tabular}{lcccccccr}

\toprule \multicolumn{1}{c}{\multirow[c]{2}{*}{Method}} & \multicolumn{1}{c}{\multirow[c]{2}{*}{Learning-rate}} & \multirow[c]{2}{*}{Rank} & AIME 2024 & MATH 500 & GPQA & AIME 2025 & \multirow[c]{2}{*}{Avg.} \\ & & & Avg@16 & Avg@4 & Avg@4 & Avg@16 & \\ \midrule

         S1 (Full-FT)          & 1e-5&- & 56.7 & 95.4 & 63.6&50 &66.43   \\
         S1 (Full-FT)          & 1e-4&- & 10&	5.9&	3.0	&0.7	&4.9  \\
        LoRA       & 1e-5   & 256 &36.67&	92.6	&58.08	&33.3&	55.16\\
        LoRA       & 1e-4   & 256 &36.67&	89.6&	52.53&	20&	49.7	\\
        Bone       &  1e-5  & 256 &43.33	&93.4	&57.58	&33.3&	56.90\\
        Bone       &  1e-4  & 256 &33.3	  &90.6	 &52.02	 &46.67	 &55.65   \\
        MoRA       & 1e-5   & 256 &50&	93.2&	62.12&	33.3& 59.66 \\
        MoRA       & 1e-4  & 256   & 26.67 &	89.4&	48.48	&33.33	&49.47 \\
          \bottomrule
        \end{tabular}
        \end{adjustbox}
      \end{sc}
    \end{small}
  \end{center}
  \vskip -0.1in
\end{table*}

\begin{table*}[t]
  \caption{Accuracy (\%) of different parameter-efficient fine-tuning methods under various adjustment styles and ranks on reasoning-intensive tasks.}
  \label{albadecay_tabledddd}
  \begin{center}
    \begin{small}
      \begin{sc}
      \begin{adjustbox}{max width=\textwidth}
        \begin{tabular}{lcccccccr}

\toprule \multicolumn{1}{c}{\multirow[c]{2}{*}{Method}} & \multicolumn{1}{c}{\multirow[c]{2}{*}{Adjustment style}} & \multirow[c]{2}{*}{Rank} & AIME 2024 & MATH 500 & GPQA & AIME 2025 & \multirow[c]{2}{*}{Avg.} \\ & & & Avg@16 & Avg@4 & Avg@4 & Avg@16 & \\ \midrule

          Full-FT        & -   & - & 56.7 & 95.4 & 63.6&50 &66.43   \\
         Full-FT     &  Stage & - &  33.33 & 	85.4 & 	46.97 & 	20 & 	46.43\\
          \midrule
          LoRA       & -   & 256 &36.67&	92.6	&58.08	&33.33&	55.16\\
          LoRA       & Linear   & 256 &53.3	&93.8	&62.63	&33.3	&60.76\\
          LoRA       & Stage   & 256 & 46.67	&93.8	&65.66&	46.67	&63.2 \\
            LoRA       & exponential   & 256  & 63.33	  & 91.4	 &63.13 &	33.33 	&62.79 \\
          LoRA       & NormTune   & 256 & 63.33&	93.6&	64.14&	53.33&	\textbf{68.6}\\
          Bone       & -   & 256 &43.33	&93.4	&57.58	&33.3&	56.90\\
          Bone       & Linear   & 256 & 50	&93.6&	55.56&	40&	59.79\\
          Bone       & Stage   & 256 &60 &92 &	61.11&	40	&63.28\\
            Bone       & Exponential    & 256 & 43.33	 &93.2	 &64.14	 &40	&60.17 \\
        Bone       & NormTune   & 256 & 56.67&	94.8&	64.65&53.33	& \textbf{67.36}\\
          MoRA       & -   & 256 &50&	93.2&	62.12&	33.3&59.66 \\
          MoRA       & Linear   & 256 &56.67& 	93& 	64.14& 	33.33& 	61.79\\
           MoRA       & Stage   & 256 &50 &94 &	59.09&	40& 	60.78 \\
             MoRA       & Exponential    & 256   & 56.67  &	92.8	  &62.63	  &40	&63.03 \\
        MoRA       & NormTune   & 256 &60&	94.6&	63.13&	60	&\textbf{69.43}\\

          \bottomrule
        \end{tabular}
        \end{adjustbox}
      \end{sc}
    \end{small}
  \end{center}
  \vskip -0.1in
\end{table*}

\begin{table*}[t]
  \caption{Accuracy (\%) of different parameter-efficient fine-tuning methods on reasoning-intensive tasks.}
  \label{s1_table}
  \begin{center}
    \begin{small}
      \begin{sc}
      \begin{adjustbox}{max width=\textwidth}
        \begin{tabular}{lcccccc}
\toprule \multicolumn{1}{c}{\multirow[c]{2}{*}{Method}}  & \multirow[c]{2}{*}{Rank} & AIME 2024 & MATH 500 & GPQA & AIME 2025 & \multirow[c]{2}{*}{Avg.} \\ &  & Avg@16 & Avg@4 & Avg@4 & Avg@16 & \\ \midrule

          Full-FT    & - & 56.7 & 95.4 & 63.6&50 &66.43   \\
          LoRA        & 8 &10&	78.8&	47.47&	13.33&	37.4\\
          Bone        & 8 &26.67&	80.4&	52.53&	40 &	49.9\\
           MoRA        & 8 &26.67	&80.4&	51.01	&20&	44.52 \\
           
          \bottomrule
        \end{tabular}
        \end{adjustbox}
      \end{sc}
    \end{small}
  \end{center}
  \vskip -0.1in
\end{table*}

\begin{table*}[t]
  \caption{Sensitivity analysis of the number of probing steps with NormTune+LoRA (rank=256).}
  \label{probing_steps_table}
  \begin{center}
    \begin{small}
      \begin{sc}
      \begin{adjustbox}{max width=\textwidth}
        \begin{tabular}{lccccccc}
        \toprule
          Method & Rank & Probing Steps & AIME 2024 & MATH 500 & GPQA & AIME 2025 & Avg. \\
        \midrule

          NormTune+LoRA (Ours) & 256 & 10 & 53.33 & 93.6 & 64.14 & 20 & 57.77 \\
          NormTune+LoRA (Ours) & 256 & 30 & 63.33 & 93.6 & 64.14 & 53.33 & 68.6 \\
          NormTune+LoRA (Ours) & 256 & 50 & 63.33 & 93.4 & 62.12 & 53.33 & 68.05 \\

        \bottomrule
        \end{tabular}
        \end{adjustbox}
      \end{sc}
    \end{small}
  \end{center}
  \vskip -0.1in
\end{table*}





\begin{table*}[t]
  \caption{Sensitivity analysis of the bucket number $n$ on Qwen3-4B using NormTune+LoRA (rank=256) on GPQA.}
  \label{bucket_number_table}
  \begin{center}
    \begin{small}
      \begin{sc}
      \begin{adjustbox}{max width=\textwidth}
        \begin{tabular}{lccc}
        \toprule
          Method & Rank & $n$ & Avg@4 \\
        \midrule

          NormTune+LoRA (Ours) & 256 & 4  & 41.3 \\
          NormTune+LoRA (Ours) & 256 & 8  & 40.9 \\
          NormTune+LoRA (Ours) & 256 & 16 & 42.6 \\
          NormTune+LoRA (Ours) & 256 & 32 & 41.7 \\

        \bottomrule
        \end{tabular}
        \end{adjustbox}
      \end{sc}
    \end{small}
  \end{center}
  \vskip -0.1in
\end{table*}

\begin{table*}[t]
  \caption{Delete? Performance of LoRA and NormTune+LoRA (rank=8) on commonsense reasoning and long-context understanding benchmarks.}
  \label{commonsense_longcontext_table}
  \begin{center}
    \begin{small}
      \begin{sc}
      \begin{adjustbox}{max width=\textwidth}
        \begin{tabular}{llc}
        \toprule
          Benchmark & Method & Avg. \\
        \midrule

          Commonsense Reasoning 
          & LoRA 
          & 90.66 \\

          Commonsense Reasoning 
          & NormTune+LoRA (Ours) 
          & 90.98 \\

          GovReport 
          & LoRA 
          & 32.44 \\

          GovReport 
          & NormTune+LoRA (Ours) 
          & 34.21 \\

        \bottomrule
        \end{tabular}
        \end{adjustbox}
      \end{sc}
    \end{small}
  \end{center}
  \vskip -0.1in
\end{table*}

\begin{table*}[t]
  \caption{Learning rate sensitivity analysis of LoRA with rank 256.}
  \label{learning_rate_table}
  \begin{center}
    \begin{small}
      \begin{sc}
      \begin{adjustbox}{max width=\textwidth}
        \begin{tabular}{lccc}
        \toprule
          Method & Rank & Learning Rate & Avg. \\
        \midrule

          LoRA & 256 & 6e-6 & 52.1 \\
          LoRA & 256 & 8e-6 & 52.9 \\
          LoRA & 256 & 1e-5 & 55.16 \\
          LoRA & 256 & 3e-5 & 53.36 \\
          LoRA & 256 & 5e-5 & 52.0 \\
          LoRA & 256 & 1e-4 & 49.7 \\

        \bottomrule
        \end{tabular}
        \end{adjustbox}
      \end{sc}
    \end{small}
  \end{center}
  \vskip -0.1in
\end{table*}

\begin{table*}[t]
  \caption{Accuracy comparison of different parameter-efficient fine-tuning methods across multiple commonsense reasoning benchmarks.}
  \label{commonsense-table}
  \begin{center}
    \begin{small}
      \begin{sc}
      \begin{adjustbox}{max width=\textwidth}
        \begin{tabular}{lccccccccccr}
          \toprule
          Method  &  Rank   & BoolQ & PIQA&SIQA & WinoGrande & ARC-e & OBQA & HellaSwag & ARC-c  & Avg.\\
          \midrule
        Full-FT        & - &  77.46 &	93.96&	81.99&	92.34&	97.73&	97.2&	96.24&	92.32&	91.16\\
        LoRA       & 8  &76.23		&92.6	&	81.06		&89.02		&98.56	&	95.8	&	96.99	&	95.05	&90.66 \\
        Bone       & 8 &73.37 &92.93&82.44&89.5&98.44& 95.6&96.56&94.96&90.48   \\
        MoRA       & 8    & 72.75& 	91.45& 	80.91& 	85.48	& 98.27& 	94.8	& 96.22& 	94.37& 	89.28  \\
          \bottomrule
        \end{tabular}
        \end{adjustbox}
      \end{sc}
    \end{small}
  \end{center}
  \vskip -0.1in
\end{table*}

\begin{table*}[t]
  \caption{ROUGE results (\%) on the GovReport dataset.}
  \label{government_table}
  \begin{center}
    \begin{small}
      \begin{sc}
        \begin{tabular}{lcccccc}
          \toprule
          Method & Rank & ROUGE-1 & ROUGE-2 & ROUGE-L & ROUGE-Lsum & Average \\
          \midrule
          Full-FT & - & 55.11 & 27.39 & 28.95 & 28.95 & 35.10 \\
          LoRA             & 8 & 52.6&	20.63&	24.67&	31.84&	32.44 \\
          Bone & 8&  60.64&	30.45&	31.94&	31.93&	38.74 \\
           MoRA & 8&  58.8&		30&		31.29&		31.08&		37.79 \\
          \bottomrule
        \end{tabular}
      \end{sc}
    \end{small}
  \end{center}
  \vskip -0.1in
\end{table*}

           

\subsubsection{Other Experiments}
\label{Other Experiments}

We provide additional results to further support the observations in the main experiments. Table~\ref{albations_table} reports the detailed results corresponding to Figure~\ref{fig:rank_analysis}. Across all ranks, NormTune consistently improves LoRA on reasoning-intensive tasks. The gains are especially pronounced at low ranks. Notably, when the rank is set to 256, NormTune+LoRA achieves an average accuracy of 68.6\%, slightly surpassing Full-FT, which further demonstrates the effectiveness of the proposed method.

Tables~\ref{s1_table}, \ref{commonsense-table}, and \ref{government_table} provide the detailed results corresponding to Figure~\ref{fig:tasks_compare}. These results highlight a clear difference between reasoning-intensive tasks and other task types. On commonsense reasoning and long-context understanding tasks, PEFT methods with rank 8 already achieve performance close to Full-FT. In contrast, the same low-rank PEFT setting performs substantially worse on reasoning-intensive tasks, as shown in Table~\ref{s1_table}.

\begin{figure}[t]
    \centering
    \includegraphics[width=0.79\linewidth]{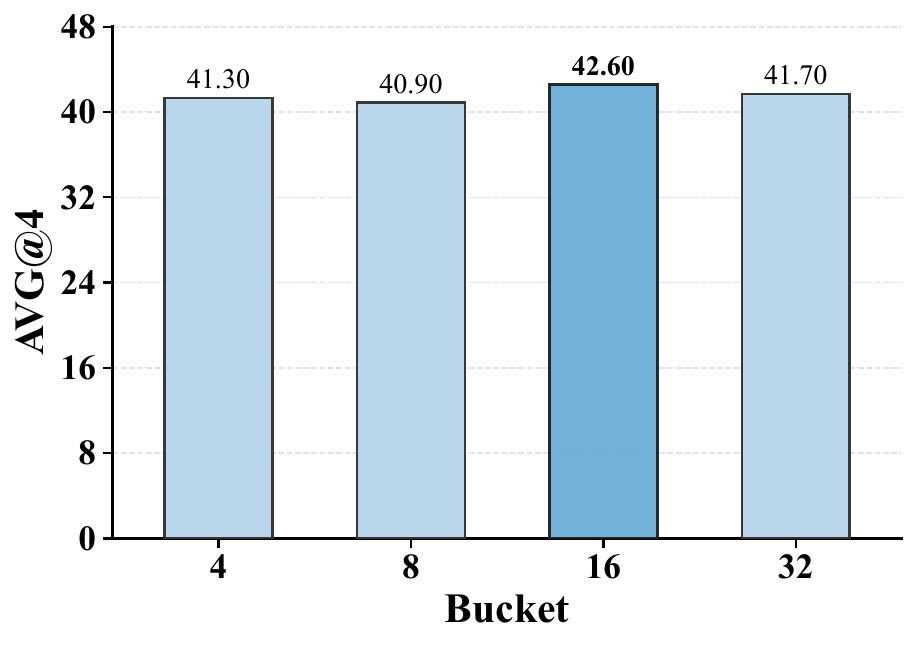}
    \caption{
        Effect of the bucket number $n$ in NormTune on GPQA after training on Qwen3-4B.
    }
    \label{fig:gpqa4b}
\end{figure}

In addition, we further conduct a sensitivity study using 10 (before warm-up), 30, and 50 (after warm-up) steps with LoRA (rank=256) under the same setting as Table ~\ref{main_table}. As shown in Table \ref{probing_steps_table}, 10 steps lead to noticeably worse performance, while the results remain stable after 30 steps, supporting our choice. Since the probing steps are included in the total training budget, using 50 steps delays the activation of NormTune by 20 updates compared to 30 steps, which may affect norm adaptation and lead to a slight performance drop. 
We also conduct an additional ablation on Qwen3-4B using NormTune+LoRA(256) on GPQA. Figure \ref{fig:gpqa4b} show the same trend: performance is relatively insensitive to 
$n$, and $n=16$ remains a strong default. This suggests that the optimal bucket number does not strongly depend on model size or depth in our tested settings. 

\begin{figure}[t]
    \centering
    \includegraphics[width=0.99\linewidth]{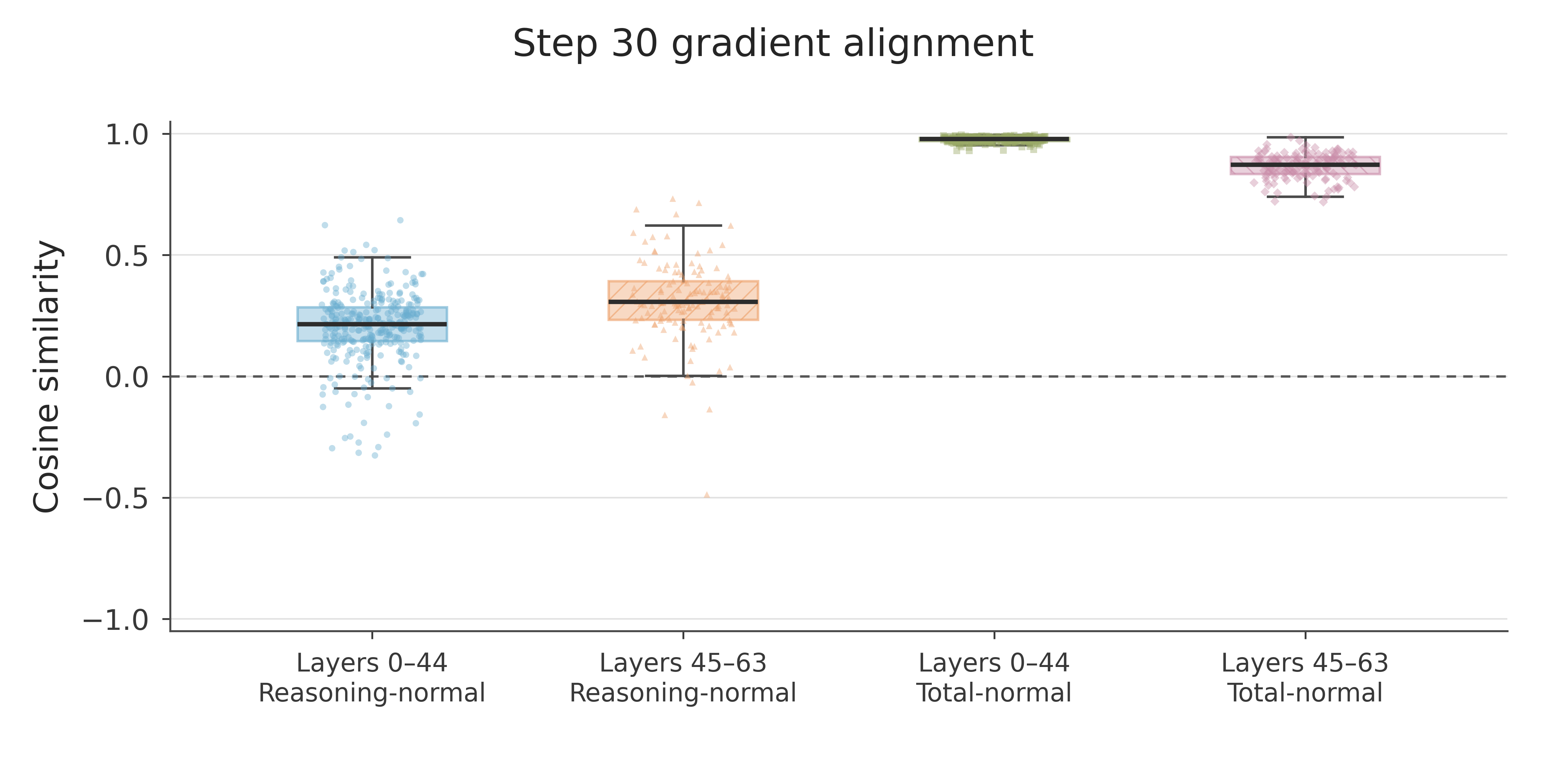}
    \caption{
        Gradient alignment across early and late layers, measured by
    $\mathrm{CosSim}(\nabla_{\theta}\mathcal{L}_{\mathrm{reasoning}},
    \nabla_{\theta}\mathcal{L}_{\mathrm{normal}})$ and
    $\mathrm{CosSim}(\nabla_{\theta}\mathcal{L}_{\mathrm{total}},
    \nabla_{\theta}\mathcal{L}_{\mathrm{normal}})$.
    }
    \label{fig:full_xiang}
\end{figure}


\subsection{Environment and Settings}
\label{Environment and Settings}

All experiments are conducted on eight NVIDIA A800 80GB GPUs. Unless otherwise specified, we follow the original training settings of S1 \cite{Muennighoff2025}, Z1 \cite{Yu2025Z1}, and Data-Repetition \cite{kopiczko2026data} for the main hyperparameters, including the warmup schedule, the number of training steps, the optimizer, and the use of gradient checkpointing. We use Qwen3-4B, Qwen-2.5-Coder-7B-Instruct\cite{Yu2025Z1}, OLMo-3-7B\cite{olmo2025olmo} and Qwen2.5-32B-Instruct as the base models for our PEFT experiments across different benchmarks. 
Unless otherwise specified, we compute layer-wise norm after training step 30, which is included in the full training steps, set the number of buckets to $n=16$, set base learning rate to the $\eta_{\min}$ of each PEFT method, set the random seed to 42 during training, and use $\eta_{\max}=1\times10^{-4}$. In addition, to ensure a fair comparison, Full-FT and PEFT use identical optimization hyperparameters, following the configurations adopted in the corresponding papers, which also include experiments on the GovReport \cite{huang2021efficient} and commonsense reasoning benchmarks. NormTune does not require a separate probing run. The first 30 updates are included in the normal training budget. The additional cost of the whole probing stage is negligible: on Qwen2.5-32B-Instruct, it requires only 0.018 GPU-hours on 8 A800 GPUs. 

Beyond S1K-1.1 \cite{Muennighoff2025}, Z1-Code-Reasoning107K \cite{Yu2025Z1}, and Data-Repetition \cite{kopiczko2026data}, 
we further fine-tune and evaluate on two additional benchmarks: commonsense reasoning and long-context understanding. For commonsense reasoning, we consider eight representative tasks, including HellaSwag \cite{Zellers_Holtzman_Bisk_Farhadi_Choi_2019}, OpenBookQA (OBQA) \cite{Mihaylov_Clark_Khot_Sabharwal_2018}, Social IQA (SIQA) \cite{Sap_Rashkin_Chen_LeBras_Choi_2019}, WinoGrande \cite{Sakaguchi_LeBras_Bhagavatula_Choi_2020}, BoolQ \cite{Clark_Lee_Chang_Kwiatkowski_Collins_Toutanova_2019}, PIQA \cite{Bisk_Zellers_Lebras_Gao_Choi_2020}, ARC-Challenge (ARC-c), and ARC-Easy (ARC-e) \cite{clark2018think}. For long-context understanding, we use GovReport \cite{huang2021efficient}, a long-document summarization benchmark consisting of reports from U.S. government agencies and their corresponding summaries. We randomly sample 1,000 examples from the training split for fine-tuning and evaluate the resulting models on the official test split.
These evaluations enable a comprehensive assessment of the effectiveness of our method. 
The detailed experimental configurations for full fine-tuning and PEFT methods, including LoRA, Bone, and MoRA, are reported in 
Tables~\ref{tab:hyperparameters},
\ref{tab:hyperparameters_adapter},
\ref{tab:hyperparameters_tempbalance},
and~\ref{tab:evaluation_settings}.
We set $a=16,512$ for RsLoRA and $exp\_temp=10$ and $esd\_update\_steps=2$ for TempBalance.
All experiments are conducted on eight NVIDIA A800 GPUs and completed within 24 hours.

\subsection{Prompts for Evaluation}
\label{Prompts for Evaluation}
For the S1K-1.1 benchmark, we adopt the prompts provided in the lm-evaluation-harness GitHub repository. 
For Z1, we adopt the prompts from the Qwen2.5-Coder GitHub repository for LiveCodeBench and BigCodeBench, while using the following prompts for MATH-500 and GPQA Diamond:
\begin{tcolorbox}[
    title=\textbf{Prompts},
    colback=white,
    colbacktitle=blue!20,
    coltitle=black,
    fonttitle=\bfseries,
    boxrule=1pt,
    arc=3mm,
    left=4mm,
    right=4mm,
    top=3mm,
    bottom=3mm
]
\texttt{<|im\_start|>system}\\
Please reason step by step, and put your final answer within
\texttt{\textbackslash boxed\{\}}.\\
\texttt{<|im\_end|>}\\
\texttt{<|im\_start|>user}\\
\texttt{\{problem\}}\\
\texttt{<|im\_end|>}\\
\texttt{<|im\_start|>assistant}
\end{tcolorbox}

For Data-Repetition, we follow the evaluation protocol in the official implementation. AIME 2024 and AIME 2025 use the same prompt, which requires the final integer answer to be enclosed in \texttt{\textbackslash boxed{}}. For GPQA Diamond, the model is instructed to place the letter corresponding to the selected answer in \texttt{\textbackslash boxed{}}. Each query is formatted using the model's default chat template. 

For commonsense reasoning, we evaluate on BoolQ, PIQA, SIQA, HellaSwag,
WinoGrande, ARC-Easy, ARC-Challenge, and OpenBookQA. For each benchmark,
the task-specific instruction is wrapped with the Alpaca-style prompt
template used in the official DoRA implementation:

\begin{tcolorbox}[
    title=\textbf{Prompts},
    colback=white,
    colbacktitle=blue!20,
    coltitle=black,
    fonttitle=\bfseries,
    boxrule=1pt,
    arc=3mm,
    left=4mm,
    right=4mm,
    top=3mm,
    bottom=3mm,
    breakable
]

Below is an instruction that describes a task. Write a response that appropriately completes the request.\par
\vspace{1mm}

\texttt{\#\#\# Instruction:}\par
\vspace{1mm}

\texttt{\detokenize{{instruction}}}\par
\vspace{1mm}

\texttt{\#\#\# Response:}

\end{tcolorbox}

For GovReport, we adopt the following prompt:
\begin{tcolorbox}[
    title=\textbf{Prompts},
    colback=white,
    colbacktitle=blue!20,
    coltitle=black,
    fonttitle=\bfseries,
    boxrule=1pt,
    arc=3mm,
    left=4mm,
    right=4mm,
    top=3mm,
    bottom=3mm,
    breakable
]

Read the following government report and write a faithful, comprehensive executive summary. Focus on the key findings, background, and conclusions. Do not invent information.\par
\vspace{2mm}

\texttt{\#\#\# Report}\par
\vspace{1mm}

\texttt{\detokenize{{report}}}\par
\vspace{2mm}

\texttt{\#\#\# Executive Summary}

\end{tcolorbox}


\begin{table*}[t]
\caption{Hyperparameter settings for GovReport and commonsense reasoning training configurations.}
\label{tab:hyperparameters_gove_comm}
\centering
\small
\begin{tabular}{lcccccccc}
\toprule
 \multirow{2}{*}{Hyperparameter} & \multicolumn{4}{c}{GovReport} & \multicolumn{4}{c}{\parbox{2.2cm}{\centering commonsense \\reasoning}} \\
\cmidrule(lr){2-5} \cmidrule(lr){6-9}
 & Full-FT & LoRA & Bone & MoRA & Full-FT & LoRA & Bone & MoRA \\
\midrule
Batch size    & 16 & 16 & 16 & 16 & 128 & 128 & 128 & 128 \\
Optimizer     & Adam & Adam & Adam & Adam & Adam & Adam & Adam & Adam \\
LR            & 1e-5 & 1e-5 & 1e-5 & 1e-5 & 3e-5 & 3e-5 & 3e-5 & 3e-5 \\
Warmup Ratio  & 0.05 & 0.05 & 0.05 & 0.05 & 0.05 & 0.05 & 0.05 & 0.05 \\
Epoch         & 5 & 5 & 5 & 5 & 3 & 3 & 3 & 3 \\
Weight Decay  & 1e-4 & 1e-4 & 1e-4 & 1e-4 & 1e-4 & 1e-4 & 1e-4 & 1e-4 \\
 Rank   & -- & 8 & 8 & 8 & -- & 8 & 8 & 8 \\
LoRA $\alpha$ & -- & 16& -- & -- & -- & 16 & --& --\\
 MoRA Type & --& -- & --& 6 & --& -- & --& 6\\
Block Size    & 32768 & 32768 & 32768 & 32768 & 256 & 256 & 256 & 256 \\
\bottomrule
\end{tabular}
\end{table*}

\begin{table*}[t]
\caption{ NormTune-specific configuration for different training benchmark. \underline{\hspace{0.2cm}} provides the optimal choice.}
\label{tab:hyperparameters}
\centering
\small
\begin{tabular}{lcccc}
\toprule
Hyperparameter & S1K-1.1 & \parbox{2.2cm}{\centering Z1-Code-\\Reasoning107K} & Data-repetition & OLMo-3-7B \\
\midrule
Batch size    & 16 & 128 & 1 & 1 \\
Optimizer     & Adam & Adam & 8-bit Adam & 8-bit Adam \\
$ \eta_{\min}$ & 1e-5 & 1e-5 & 1e-5, \underline{3e-5} & 2e-5 \\
$ \eta_{\max}$     & {\underline{1e-4},1e-3} & 2e-5,\underline{1e-4} &2e-5,3e-5,\underline{1e-4} &1e-4 \\
$n$ buckets& 16& 16& 16& 16\\
Probing Steps & 30& 30&30&30\\
Warmup Ratio & 0.05 & 0.06 & 0.1 & 0.1 \\
Epoch       & 5 & 2 & 16 & 16 \\
Weight Decay       & 1e-4 & 0.0 & 0.0 & 0.0 \\
Target Module & \multicolumn{4}{c}{Q, K, V, O, Gate, Up, Down} \\
Block Size & 32768 &8192 &32768 &32768 \\
\bottomrule
\end{tabular}
\end{table*}

\begin{table*}[t]
\caption{Hyperparameter settings for different PEFT training configurations.}
\label{tab:hyperparameters_adapter}
\centering
\small
\setlength{\tabcolsep}{3pt}
\resizebox{\textwidth}{!}{
\begin{tabular}{llccccccccccc}
\toprule
Configuration & Method
& Batch Size
& Optimizer
& LR
& Warmup Ratio
& Epoch
& Weight Decay
& Rank
& LoRA $\alpha$
& MoRA Type
& AdaLoRA\_init\_rank
& Block Size \\
\midrule
\multirow{9}{*}{S1K-1.1}
& Full-FT     & 16 & Adam & 1e-5 & 0.05 & 5 & 1e-4 & -- & --  & -- & -- & 32768 \\
& LoRA        & 16 & Adam & 1e-5 & 0.05 & 5 & 1e-4 & 8,256 & 16,512 & -- & -- & 32768 \\
& Bone        & 16 & Adam & 1e-5 & 0.05 & 5 & 1e-4 & 8,256 & --  & -- & -- & 32768 \\
& MoRA        & 16 & Adam & 1e-5 & 0.05 & 5 & 1e-4 & 8,256 & --  & 6,1  & -- & 32768 \\
& AdaLoRA     & 16 & Adam & 1e-5 & 0.05 & 5 & 1e-4 & 8,256 & --  & -- & 12,384 & 32768 \\
& DoRA        & 16 & Adam & 1e-5 & 0.05 & 5 & 1e-4 & 8,256 & --  & -- & -- & 32768 \\
& RSLoRA      & 16 & Adam & 1e-5 & 0.05 & 5 & 1e-4 & 8,256 & --  & -- & -- & 32768 \\
& LoRA+       & 16 & Adam & 1e-5 & 0.05 & 5 & 1e-4 & 8,256 & --  & -- & -- & 32768 \\
& TempBalance & 16 & Adam & 1e-5 & 0.05 & 5 & 1e-4 & 8,256 & --  & -- & -- & 32768 \\
\midrule
\multirow{9}{*}{\parbox{2.2cm}{\centering Z1-Code-\\Reasoning107K}}
& Full-FT     & 128 & Adam & 1e-5 & 0.06 & 2 & 0.0 & 256 & --  & -- & -- & 8192 \\
& LoRA        & 128 & Adam & 1e-5 & 0.06 & 2 & 0.0 & 256 & 512 & -- & -- & 8192 \\
& Bone        & 128 & Adam & 1e-5 & 0.06 & 2 & 0.0 & 256 & --  & -- & -- & 8192 \\
& MoRA        & 128 & Adam & 1e-5 & 0.06 & 2 & 0.0 & 256 & --  & 1  & -- & 8192 \\
& AdaLoRA     & 128 & Adam & 1e-5 & 0.06 & 2 & 0.0 & 256 & --  & -- & 384 & 8192 \\
& DoRA        & 128 & Adam & 1e-5 & 0.06 & 2 & 0.0 & 256 & --  & -- & -- & 8192 \\
& RSLoRA      & 128 & Adam & 1e-5 & 0.06 & 2 & 0.0 & 256 & --  & -- & -- & 8192 \\
& LoRA+       & 128 & Adam & 1e-5 & 0.06 & 2 & 0.0 & 256 & --  & -- & -- & 8192 \\
& TempBalance & 128 & Adam & 1e-5 & 0.06 & 2 & 0.0 & 256 & --  & -- & -- & 8192 \\
\midrule
\multirow{9}{*}{Data-repetition}
& Full-FT     & 1 & 8-bit Adam & 3e-5 & 0.1 & 16 & 0.0 & 256 & --  & -- & -- & 32768 \\
& LoRA        & 1 & 8-bit Adam & 3e-5 & 0.1 & 16 & 0.0 & 256 & 512 & -- & -- & 32768 \\
& Bone        & 1 & 8-bit Adam & 3e-5 & 0.1 & 16 & 0.0 & 256 & --  & -- & -- & 32768 \\
& MoRA        & 1 & 8-bit Adam & 3e-5 & 0.1 & 16 & 0.0 & 256 & --  & 1  & -- & 32768 \\
& AdaLoRA     & 1 & 8-bit Adam & 3e-5 & 0.1 & 16 & 0.0 & 256 & --  & -- & 384 & 32768 \\
& DoRA        & 1 & 8-bit Adam & 3e-5 & 0.1 & 16 & 0.0 & 256 & --  & -- & -- & 32768 \\
& RSLoRA      & 1 & 8-bit Adam & 3e-5 & 0.1 & 16 & 0.0 & 256 & --  & -- & -- & 32768 \\
& LoRA+       & 1 & 8-bit Adam & 3e-5 & 0.1 & 16 & 0.0 & 256 & --  & -- & -- & 32768 \\
& TempBalance & 1 & 8-bit Adam & 3e-5 & 0.1 & 16 & 0.0 & 256 & --  & -- & -- & 32768 \\
\midrule
\multirow{9}{*}{OLMo-3-7B}
& Full-FT     & 1 & 8-bit Adam & 2e-5 & 0.1 & 16 & 0.0 & 256 & --  & -- & -- & 32768 \\
& LoRA        & 1 & 8-bit Adam & 2e-5 & 0.1 & 16 & 0.0 & 256 & 512 & -- & -- & 32768 \\
& Bone        & 1 & 8-bit Adam & 2e-5 & 0.1 & 16 & 0.0 & 256 & --  & -- & -- & 32768 \\
& MoRA        & 1 & 8-bit Adam & 2e-5 & 0.1 & 16 & 0.0 & 256 & --  & 1  & -- & 32768 \\
& AdaLoRA     & 1 & 8-bit Adam & 2e-5 & 0.1 & 16 & 0.0 & 256 & --  & -- & 384 & 32768 \\
& DoRA        & 1 & 8-bit Adam & 2e-5 & 0.1 & 16 & 0.0 & 256 & --  & -- & -- & 32768 \\
& RSLoRA      & 1 & 8-bit Adam & 2e-5 & 0.1 & 16 & 0.0 & 256 & --  & -- & -- & 32768 \\
& LoRA+       & 1 & 8-bit Adam & 2e-5 & 0.1 & 16 & 0.0 & 256 & --  & -- & -- & 32768 \\
& TempBalance & 1 & 8-bit Adam & 2e-5 & 0.1 & 16 & 0.0 & 256 & --  & -- & -- & 32768 \\
\bottomrule
\end{tabular}
}
\end{table*}

\begin{table*}[t]
\caption{Hyperparameter settings for TempBalance configurations.}
\label{tab:hyperparameters_tempbalance}
\centering
\small
\setlength{\tabcolsep}{3pt}
\resizebox{\textwidth}{!}{
\begin{tabular}{llccccccccccc}
\toprule
Configuration & Method
& Batch Size
& Optimizer
& LR
& Warmup Ratio
& Epoch
& Weight Decay
& Rank
& LoRA $\alpha$
& MoRA Type
& AdaLoRA\_init\_rank
& Block Size \\
\midrule
\multirow{9}{*}{S1K-1.1}
& Full-FT     & 16 & Adam & 1e-5 & 0.05 & 5 & 1e-4 & -- & --  & -- & -- & 32768 \\
& LoRA        & 16 & Adam & 1e-5 & 0.05 & 5 & 1e-4 & 8,256 & 16,512 & -- & -- & 32768 \\
& Bone        & 16 & Adam & 1e-5 & 0.05 & 5 & 1e-4 & 8,256 & --  & -- & -- & 32768 \\
& MoRA        & 16 & Adam & 1e-5 & 0.05 & 5 & 1e-4 & 8,256 & --  & 6,1  & -- & 32768 \\
& AdaLoRA     & 16 & Adam & 1e-5 & 0.05 & 5 & 1e-4 & 8,256 & --  & -- & 12,384 & 32768 \\
& DoRA        & 16 & Adam & 1e-5 & 0.05 & 5 & 1e-4 & 8,256 & --  & -- & -- & 32768 \\
& RSLoRA      & 16 & Adam & 1e-5 & 0.05 & 5 & 1e-4 & 8,256 & --  & -- & -- & 32768 \\
& LoRA+       & 16 & Adam & 1e-5 & 0.05 & 5 & 1e-4 & 8,256 & --  & -- & -- & 32768 \\
& TempBalance & 16 & Adam & 1e-5 & 0.05 & 5 & 1e-4 & 8,256 & --  & -- & -- & 32768 \\
\midrule
\multirow{9}{*}{\parbox{2.2cm}{\centering Z1-Code-\\Reasoning107K}}
& Full-FT     & 128 & Adam & 1e-5 & 0.06 & 2 & 0.0 & 256 & --  & -- & -- & 8192 \\
& LoRA        & 128 & Adam & 1e-5 & 0.06 & 2 & 0.0 & 256 & 512 & -- & -- & 8192 \\
& Bone        & 128 & Adam & 1e-5 & 0.06 & 2 & 0.0 & 256 & --  & -- & -- & 8192 \\
& MoRA        & 128 & Adam & 1e-5 & 0.06 & 2 & 0.0 & 256 & --  & 1  & -- & 8192 \\
& AdaLoRA     & 128 & Adam & 1e-5 & 0.06 & 2 & 0.0 & 256 & --  & -- & 384 & 8192 \\
& DoRA        & 128 & Adam & 1e-5 & 0.06 & 2 & 0.0 & 256 & --  & -- & -- & 8192 \\
& RSLoRA      & 128 & Adam & 1e-5 & 0.06 & 2 & 0.0 & 256 & --  & -- & -- & 8192 \\
& LoRA+       & 128 & Adam & 1e-5 & 0.06 & 2 & 0.0 & 256 & --  & -- & -- & 8192 \\
& TempBalance & 128 & Adam & 1e-5 & 0.06 & 2 & 0.0 & 256 & --  & -- & -- & 8192 \\
\midrule
\multirow{9}{*}{Data-repetition}
& Full-FT     & 1 & 8-bit Adam & 3e-5 & 0.1 & 16 & 0.0 & 256 & --  & -- & -- & 32768 \\
& LoRA        & 1 & 8-bit Adam & 3e-5 & 0.1 & 16 & 0.0 & 256 & 512 & -- & -- & 32768 \\
& Bone        & 1 & 8-bit Adam & 3e-5 & 0.1 & 16 & 0.0 & 256 & --  & -- & -- & 32768 \\
& MoRA        & 1 & 8-bit Adam & 3e-5 & 0.1 & 16 & 0.0 & 256 & --  & 1  & -- & 32768 \\
& AdaLoRA     & 1 & 8-bit Adam & 3e-5 & 0.1 & 16 & 0.0 & 256 & --  & -- & 384 & 32768 \\
& DoRA        & 1 & 8-bit Adam & 3e-5 & 0.1 & 16 & 0.0 & 256 & --  & -- & -- & 32768 \\
& RSLoRA      & 1 & 8-bit Adam & 3e-5 & 0.1 & 16 & 0.0 & 256 & --  & -- & -- & 32768 \\
& LoRA+       & 1 & 8-bit Adam & 3e-5 & 0.1 & 16 & 0.0 & 256 & --  & -- & -- & 32768 \\
& TempBalance & 1 & 8-bit Adam & 3e-5 & 0.1 & 16 & 0.0 & 256 & --  & -- & -- & 32768 \\
\midrule
\multirow{9}{*}{OLMo-3-7B}
& Full-FT     & 1 & 8-bit Adam & 2e-5 & 0.1 & 16 & 0.0 & 256 & --  & -- & -- & 32768 \\
& LoRA        & 1 & 8-bit Adam & 2e-5 & 0.1 & 16 & 0.0 & 256 & 512 & -- & -- & 32768 \\
& Bone        & 1 & 8-bit Adam & 2e-5 & 0.1 & 16 & 0.0 & 256 & --  & -- & -- & 32768 \\
& MoRA        & 1 & 8-bit Adam & 2e-5 & 0.1 & 16 & 0.0 & 256 & --  & 1  & -- & 32768 \\
& AdaLoRA     & 1 & 8-bit Adam & 2e-5 & 0.1 & 16 & 0.0 & 256 & --  & -- & 384 & 32768 \\
& DoRA        & 1 & 8-bit Adam & 2e-5 & 0.1 & 16 & 0.0 & 256 & --  & -- & -- & 32768 \\
& RSLoRA      & 1 & 8-bit Adam & 2e-5 & 0.1 & 16 & 0.0 & 256 & --  & -- & -- & 32768 \\
& LoRA+       & 1 & 8-bit Adam & 2e-5 & 0.1 & 16 & 0.0 & 256 & --  & -- & -- & 32768 \\
& TempBalance & 1 & 8-bit Adam & 2e-5 & 0.1 & 16 & 0.0 & 256 & --  & -- & -- & 32768 \\
\bottomrule
\end{tabular}
}
\end{table*}











\begin{table*}[t]
\caption{
Evaluation settings for different training configurations.
For S1K-1.1 and Z1-Code-Reasoning107K, mean greedy accuracy
is averaged over repeated evaluation runs of the same checkpoint.
}
\label{tab:evaluation_settings}
\centering
\small
\setlength{\tabcolsep}{3.5pt}
\renewcommand{\arraystretch}{1.15}

\resizebox{\textwidth}{!}{%
\begin{tabular}{
    >{\raggedright\arraybackslash}p{2.2cm}
    >{\raggedright\arraybackslash}p{3.1cm}
    >{\raggedright\arraybackslash}p{3.3cm}
    >{\raggedright\arraybackslash}p{3.8cm}
    c
    c
    >{\raggedright\arraybackslash}p{2.7cm}
}
\toprule
Configuration
& Evaluation suites
& Metric / evaluator
& Decoding
& \shortstack{Max\\tokens}
& \shortstack{Completions per\\problem per run}
& \shortstack[l]{Greedy evaluation\\runs ($R$)}
\\
\midrule

S1K-1.1
& AIME 2024, AIME 2025, MATH 500, GPQA
& Mean greedy accuracy
  (GPT-4o-mini)
& Greedy, $T=0$
& 32768
& 1
& 4: MATH 500, GPQA;
  16: AIME 2024, AIME 2025
\\
\addlinespace[3pt]

Z1-Code-\newline Reasoning107K
& LiveCodeBench, MATH 500, BigCodeBench-Hard, GPQA
& Mean greedy accuracy
  (Qwen2.5-32B-Instruct)
& Greedy, $T=0$
& 4096
& 1
& 4: MATH 500, GPQA;
  16: LiveCodeBench, BigCodeBench-Hard
\\
\addlinespace[3pt]

Data-repetition
& GPQA, AIME 2024, AIME 2025
& Avg@$k$ and Pass@$k$
& Sampling, $T=0.6$,
  top-$p=0.95$, top-$k=20$
& 32768
& \shortstack{4 (GPQA)\\16 (AIME)}
& ---
\\
\addlinespace[3pt]

OLMo-3-7B
& GPQA, AIME 2024, AIME 2025
& Avg@$k$ and Pass@$k$
& Sampling, $T=0.6$,
  top-$p=0.95$, top-$k=50$
& 32768
& \shortstack{4 (GPQA)\\16 (AIME)}
& ---
\\
\addlinespace[3pt]

GovReport
& GovReport test split
& ROUGE-1, ROUGE-2, ROUGE-L, ROUGE-Lsum
& Sampling, $T=0.1$, top-$p=0.9$
& 32768
& 1
& ---
\\
\addlinespace[3pt]

Commonsense\newline reasoning
& BoolQ, PIQA, SIQA, WinoGrande,
  ARC-e/c, OBQA, HellaSwag
& Accuracy
& 4-beam search;
  $T=0.1$, top-$p=0.75$, top-$k=40$
  in the generation configuration
& 32
& 1
& ---
\\

\bottomrule
\end{tabular}%
}

\vspace{3pt}

\begin{minipage}{\textwidth}
\footnotesize
\textit{Note.}
For S1K-1.1 and Z1-Code-Reasoning107K, each greedy evaluation
run generates one completion per problem. The reported accuracy
is the arithmetic mean of the accuracies from $R$ evaluation
runs using the same checkpoint and benchmark questions.
These runs are not independent training runs or stochastic
decoding samples.
For Data-repetition and OLMo-3-7B, $k$ denotes the number of
sampled completions per problem used to compute Avg@$k$
and Pass@$k$.
A dash in the last column indicates that the repeated-greedy
evaluation protocol does not apply.
\end{minipage}

\end{table*}

\end{document}